\documentclass{article}
\usepackage{ijcai19}

\usepackage{times}
\usepackage{soul}
\usepackage{url}
\usepackage[utf8]{inputenc}
\usepackage[small]{caption}
\usepackage{graphicx}
\usepackage{amsmath}
\usepackage{amsfonts}
\usepackage{booktabs}
\usepackage{algorithm}
\usepackage{algorithmic}
\usepackage{epstopdf-base}
\usepackage{subcaption}
\newtheorem{assumption}{Assumption}
\newtheorem{lemma}{Lemma}
\newtheorem{theorem}{Theorem}
\newtheorem{remark}{Remark}

\title{Scalable Semi-Supervised SVM via Triply Stochastic Gradients}

\author{
	Xiang Geng$^{1}$
	\and
	Bin Gu$^{1,2}$\and
	Xiang Li$^{4}$\and
	Wanli Shi$^{1}$\and
	Guansheng Zheng$^{1}$\And
	Heng Huang$^{2,3}$
	\thanks{To whom all correspondence should be addressed.}
	\affiliations
	$^{1}$School of Computer \& Software, Nanjing University of Information Science \& Technology, P.R.China\\
	$^{2}$JD Finance America Corporation \\
	$^{3}$Department of Electrical \& Computer Engineering, University of Pittsburgh, USA\\
	$^{4}$Computer Science Department, University of Western Ontario, Canada \quad
	\emails
	gengxiang@nuist.edu.cn,
	jsgubin@gmail.com,
	lxiang2@uwo.ca,
	wanlishi@nuist.edu.cn,
	zgs@nuist.edu.cn,
	heng.huang@pitt.edu
}

\begin{document}
	
\maketitle
\begin{abstract}
	Semi-supervised learning (SSL) plays an increasingly important role in the big data era because a large number of unlabeled samples can be used effectively to improve the performance of the classifier. Semi-supervised support vector machine (S$^3$VM) is one of the most appealing methods for SSL, but scaling up S$^3$VM for kernel learning is still an open problem. Recently, a doubly stochastic gradient (DSG) algorithm has been proposed to achieve efficient and scalable training for kernel methods. However, the algorithm and theoretical analysis of DSG are developed based on the convexity assumption which makes them incompetent for non-convex problems such as S$^3$VM. To address this problem, in this paper, we propose a triply stochastic gradient algorithm for S$^3$VM, called TSGS$^3$VM. Specifically, to handle two types of data instances involved in S$^3$VM, TSGS$^3$VM samples a labeled instance and  an unlabeled instance as well with the random features in each iteration to compute a triply stochastic gradient. We use the approximated gradient to update the  solution. More importantly, we establish new theoretic analysis for TSGS$^3$VM which guarantees that TSGS$^3$VM can converge to a stationary point. Extensive experimental results on a variety of datasets demonstrate that TSGS$^3$VM is much more efficient and scalable than existing S$^3$VM algorithms.
\end{abstract}

\section{Introduction}

\begin{table*}[htbp]
	\begin{center}
		\setlength{\tabcolsep}{3mm}
		\begin{tabular}{ccccc}
			\toprule
			\textbf{Algorithm}  &  \textbf{Reference}& \textbf{Method}	     &\textbf{Computational Complexity}&   \textbf{Space Complexity}\\
			\midrule
			S$^3$VM$^{light}$&\cite{joachims1999transductive}		&Self-labeling heuristics&		
			$O(tn^\kappa)$&		$O(n^2)$\\
			NTS$^3$VM&\cite{chapelle2007training} 	&Gradient-based
			&$O(n^3)$	&$O(n^2)$\\
			BGS$^3$VM&\cite{le2016budgeted} 	&Gradient-based
			&$O(n^3)$	&$O(n^2)$\\
			BLS$^3$VM& \cite{collobert2006large}		&CCCP-based&			$O(tn^\kappa)$& 		$O(n^2)$\\
			ILS$^3$VM&\cite{gu2018new}	&CCCP-based	&$\approx O(Tn^2)$	&$O(n^2)$\\
			\midrule
			TSGS$^3$VM&Our		&TSG&		$O(mT^2)$& 	 	$O(T)$\\
			\bottomrule
		\end{tabular}
	\end{center}
	\caption{Comparisons of computational complexities and memory requirements of representative S$^3$VM algorithms. ($n$ is the training size, $T$ is the total number of iteration, $t$ denotes the number of outer loops and $1<\kappa<2.3$)
		\label{tab:algorithm}}
\end{table*}

Semi-supervised learning (SSL) plays an increasingly important role in the big data era because a large number of unlabeled samples can be used effectively to improve the performance of the classifier.
Semi-supervised support vector machine (S$^3$VM) \cite{bennett1999semi} is one of the most appealing methods for SSL.
Specifically, S$^3$VM enforces the classification boundary to go across the less-dense regions in the reproducing kernel Hilbert space (RKHS), while keeping the labeled data correctly classified.  Unfortunately, this will lead to a non-convex optimization problem. It is well known that solving a non-convex optimization problem is normally  difficult than solving a convex one like standard support vector machine.
%
Under this arduous challenge, a wide spectrum of methods for S$^3$VM have been proposed in the last two decades.
Generally speaking, these methods can be roughly divided into three groups, \emph{i.e.},
methods with self-labeling heuristics, concave-convex procedure (CCCP) methods and gradient-based methods.
We give a brief review of these representative S$^3$VM methods in Section \ref{sec:related works} and Table \ref{tab:algorithm}.

Unfortunately, these traditional S$^3$VM methods are inefficient due to increased complexity introduced by the cost of kernel computation as well as non-convexity.
Specifically, the kernel matrix needs $O(n^2d)$ operations to be calculated and $O(n^2)$ memory to be stored, where $n$ denotes the size of training samples and $d$ denotes dimension of the data \cite{gu2018accelerated}. Essentially, gradient-based S$^3$VM methods have $O(n^3)$ complexity due mainly to the operations on the kernel matrix. Even though a convex kernel problem can be solved by a 
state-of-the-art solver (\emph{e.g.} LIBSVM), $O(n^\kappa)$ computation is still needed where $1<\kappa<2.3$ \cite{CC01a}.
While to handle the non-convexity of S$^3$VM, the methods using self-labeling heuristics and CCCP-based algorithms need to solve multiple convex sub-problems to guarantee that they finally converge \cite{yuille2002concave}. As a result, these methods scale as $O(tn^{\kappa})$, where $t$ denotes the number of solving sub-problems.
We summarize the computational complexities and memory requirements of the representative S$^3$VM methods in Table \ref{tab:algorithm}. 
As pointed in \cite{gu2018new}, scaling up S$^3$VM is still an open problem.

Recently, a novel doubly stochastic gradient (DSG)  method \cite{dai2014scalable} was proposed to achieve efficient and scalable training for kernel methods.
Specifically, in each iteration, DSG computes a doubly stochastic gradient by sampling a random data sample and  the corresponding random features to update the solution. Thus, DSG avoids computing and storing a kernel matrix, while enjoying   nice computational and space complexities.  Xie et al. \shortcite{xie2015scale} used  DSG to scale up nonlinear component analysis. To the best of our knowledge, \cite{xie2015scale} is the only work based on DSG to solve  a non-convex problem.

However, existing algorithms and theoretical analysis of DSG cannot be applied to S$^3$VM due to the following two reasons.
\textbf{1) Multiple data distributions:} S$^3$VM  minimizes  the training errors coming from two different sources. One is the expected error on the unlabeled data, and the other one is the  mean  error on the labeled data whose size is normally significantly smaller than the one of unlabeled data.  However, DSG only considers the expected error on the labeled data.
\textbf{2) Non-convexity analysis:} The theoretical analysis in \cite{xie2015scale} is based on a strong assumption (\emph{i.e.},  the initialization needs to be close to the optimum). However, such an  assumption is rarely satisfied in practice. Besides, they focus on the nonlinear component analysis instead of general non-convex problems. Thus, it is infeasible to extend the analysis of \cite{xie2015scale} to  S$^3$VM.


To address this challenging problem, we first propose a new and practical formulation of S$^3$VM.
Then, we develop a new triply stochastic gradient algorithm (TSGS$^3$VM) to solve the corresponding optimization problem. Specifically, to handle two types of data instances involved in S$^3$VM, TSGS$^3$VM samples a labeled instance and an unlabeled instance as well with their random features in each iteration to compute a triply stochastic gradient (TSG). We then use the TSGs to iteratively update the solution. A critical question is whether and how fast this optimization process with multiple randomness  would converge.
In addressing this concern, we establish new theoretic analysis for TSGS$^3$VM which guarantees that TSGS$^3$VM can converge to a stationary point with a sublinear convergence rate for a general non-convex learning problem under weak assumptions. Extensive experimental results  demonstrate the superiority of TSGS$^3$VM.

\noindent \textbf{Novelties.} We summary the main novelties of this paper as follows.
\begin{enumerate}
	\item To scale up S$^3$VM, we propose a practical formulation of S$^3$VM and develop a novel extension of DSG that could solve optimization problems with multiple data sources.
	\item We have established the new theoretic analysis of TSGS$^3$VM algorithm for a general non-convex learning problem which guarantees its convergence to a stationary point. To the best of our knowledge, it is the first work offering non-convex analysis for DSG-like algorithms without initialization assumption.
\end{enumerate}

\section{Related Works}\label{sec:related works}
We give a brief review of  kernel approximation methods as well as the representative S$^3$VM methods.

\paragraph{Kernel Approximation.}
There are many kernel approximation methods proposed to address the scalability issue of kernel methods.
For instance, low-rank factors are used to approximate the kernel matrix in
\cite{drineas2005nystrom}.
Rahimi \& Recht \shortcite{rahimi2008random} provided another method that uses random features to approximate the map function explicitly.
However, as analyzed in \cite{drineas2005nystrom,lopez2014randomized}, the rank for low-rank and the number of random features need to be $O(n)$ to obtain a good generalization ability.
To further improve the random features method, Dai et al. \shortcite{dai2014scalable} proposed DSG descent algorithm.
Carratino et al. \shortcite{carratino2018learning} proved that DSG only need $O(\sqrt{n})$ random features to obtain a good result.
However, existing DSG methods \cite{li2017triply,gu2018asynchronous,shi2019quadruply} can not be used for S$^3$VM as discussed previously.

\paragraph{S${^3}$VM Methods.}
As mentioned above, traditional S${^3}$VM methods can be roughly divided into three types, \emph{i.e.}, the method of self-labeling heuristics, the concave-convex procedure (CCCP) method, and the gradient-based method.
For the  method of self-labeling heuristics,
Joachims \shortcite{joachims1999transductive} proposed a S$^3$VM$^{light}$ algorithm which  uses self-labeling heuristics for labeling the unlabeled data, then iteratively solve this standard SVM until convergence.
CCCP-based methods were proposed to solve S$^3$VM in \cite{chapelle2005semi,wang2007transductive,yu2019tackle}.
The basic principle of CCCP is to linearize the concave part of S$^3$VM's objective function around a solution obtained in the current iteration so that sub-problem is convex. Then the CCCP framework solves a sequence of the convex sub-problem iteratively until decision variable converges.
Based on CCCP framework, Gu et al. \cite{gu2018new} proposed an incremental
learning method for S$^3$VM which is suitable for the online scenario.
For gradient-based methods, 
Chapelle and Zien \shortcite{chapelle2005semi} approximate the kernel matrix $K$ using low-rank factors, then using gradient descent to solve S$^3$VM on the low-rank matrix.
BGS$^3$VM \cite{le2016budgeted} uses budgeted SGD to limit the model size to two predefined budgets $B_l$ and $B_u$.

\section{Preliminaries}
In this section, we first give a general non-convex learning problem for S$^3$VM, and then give a brief review of random feature approximation.

\begin{table}[!ht]
	\setlength{\tabcolsep}{3mm}
	\begin{center}
		\begin{small}
			\begin{tabular}{ccc}
				\toprule
				Name&$u(r)$ &$u'(r)$\\
				\midrule
				SHG&$\max\{0,1-|r|\}$&$\left\{\begin{matrix}
				0 & {\rm if} \ |r|\geq 1
				\\
				-1 & {\rm if}\ |r|<1
				\end{matrix}\right. $\\
				SSHG&$\frac{1}{2}\max\{0,1-|r|\}^2$ &$\left\{\begin{matrix}
				0 & {\rm if}\ |r|\geq 1
				\\
				|r|-1 & {\rm if}\ |r|<1
				\end{matrix}\right. $\\
				Ramp&$H_{1}(r)-H_{s}(r)$ &$H'_{1}(r)-H'_{s}(r)$\\
				
				DA&$\exp(-5r^2)$ &$-10r \cdot \exp(-5r^2)$\\
				\bottomrule
			\end{tabular}
		\end{small}
	\end{center}
	\caption{Summary of the non-convex loss functions used in S$^3$VM, where $H_s(\cdot)=\max\{0,s-\cdot\}$, then $H'_{s}(\cdot)=0$, if $ \cdot \geq s$ else $H'_{s}(\cdot)=-1$. SHG, SSHG and DA denote symmetric hinge,  square SHG and  a differentiable approximation to SSHG respectively.}
	\label{tab:loss}
\end{table}
\subsection{S$^3$VM Optimization Formulation}
Given the training dataset $\cal X$ constituted with $n^l$ labeled examples $\mathcal{L}:=\{({x}_i,y_i)\}_{i=1}^{n^l}$ and $n^u$ unlabeled examples $\mathcal{U} := \{{x}_i\}_{i=n^l+1}^{n}$, where $n = n^l + n^u$, ${x_i} \in \mathbb{R}^d$, and $y_i \in \{1,-1\}$.
Traditional S$^3$VM solves the following problem.
\begin{align}
\min_{f\in \cal H} \quad &\frac{1}{2} ||f||^2_{\cal H}+\frac{C}{n^l}\sum_{(x,y)\in \mathcal{L}}l( f(x),y)
+\frac{C^*}{n^u} \sum_{x\in \mathcal{U}}u(f(x)) \nonumber
\end{align}
where $C$ and $C^*$ are regularization parameters, $||\cdot||_{\cal H}$ denotes the norm in RKHS, $l(r,v) = \max(0, 1 - vr)$ is the hinge loss, its subgradient $l'(r,v)=0$, if $rv\ge1$, else $l'(r,v)=-v$,  $u(r)$ is the non-convex loss function which enforce unlabeled data away from the discrimination hyperplane. We summarize the commonly used non-convex S$^3$VM losses and its subgradient $u'(r)$ in Table \ref{tab:loss}.

For S$^3$VM problems, however, the volumes of labeled and unlabeled data are usually quite different.
Because of the labeling cost, the labeled dataset is often very small, while a large amount of unlabeled data can be obtained relatively easily. Taking this into consideration, we propose to solve a novel S$^3$VM formulation as follows.
\begin{eqnarray}\label{S3VM2}
&&\min_{f\in \cal H} R(f)
\\
&=& \frac{1}{2} ||f||^2_{\cal H}+\frac{C}{n^l}\sum_{(x,y)\in \mathcal{L}}l( f(x),y)
+C^* \mathbb{E}_{x \sim P(x)} u(f(x)) \nonumber
\end{eqnarray}
where $P(x)$ denotes the target data distribution.
Notice that we use the empirical mean  error on the labeled dataset, while using the expected error on the whole distribution for the unlabeled data.

\subsection{Random Feature Approximation}

Random feature is a powerful technique to make kernel methods scalable. It uses the intriguing duality between kernels and stochastic processes.
Specifically, according to the Bochner theorem \cite{wendland2004scattered},
for any positive definite PD kernel $k(\cdot,\cdot)$, there exists a set $\Omega$, a probability measure $\mathbb{P}$ and a random feature map $\phi_\omega(x)$, such that
$k(x,x')=\int_{\Omega}\phi_\omega(x) \phi_\omega(x')d\mathbb{P}(\omega)$.
In this way, the value of the kernel function can be approximated by explicitly computing random features $\phi_\omega(x)=[\frac{1}{\sqrt{m}}\phi_{\omega_1}(x),\frac{1}{\sqrt{m}}\phi_{\omega_2}(x),\cdots,\frac{1}{\sqrt{m}}\phi_{\omega_m}(x)]$, \emph{i.e.},
\begin{align}\label{random features}
k(x,x')\approx \frac{1}{m}\sum_{i=1}^{m}\phi_{\omega_i}(x) \phi_{\omega_i}(x')
\end{align}
where $m$ is the number of random features.
Using Gaussian RBF kernel
as a concrete example, it yields a Gaussian distribution $\mathbb{P}(\omega)$ over random feature maps of Fourier basis functions $\phi_{\omega_i}(x) = \sqrt{2}cos({\omega_i^Tx+b})$ to compute its feature mapping, where $\omega_i$ is drawn from $\mathbb{P}(\omega)$ and $b$ is drawn uniformly form $[0,2\pi]$.
Moreover, many random feature construction methods have been proposed for various kernels, such as dot-product kernels and Laplacian kernels.

The theory of RKHS provides a rigorous mathematical framework for studying optimization problems in the functional space.
Specifically, we know that every PD kernel $k(x,x')$ has a corresponding RKHS $\mathcal{H}$. An RKHS $\mathcal{H}$
has the reproducing property, \emph{i.e.},
$\forall x\in \mathcal{X}, \forall f\in \mathcal{H}$, we always have $\langle f(\cdot),k(x,\cdot)\rangle_{\mathcal{H}}=f(x)$.
Besides, functional gradient in RKHS $\mathcal{H}$ can be computed as $\nabla f(x)= k(x, \cdot)$ and $\nabla ||f||^2_{\mathcal{H}} = 2f$.
%

\section{Triply Stochastic S$^3$VM}
The above section has introduced the basic theoretic tools for triply stochastic functional gradient descent. Now we introduce how to utilize these tools to solve the S$^3$VM problem.
\subsection{Triply Stochastic Gradient}
From Eq. (\ref{S3VM2}), it is not hard to notice that $R(f)$ involves two different data sources. Taking into consideration the distribution of random features $\omega\sim \mathbb{P}(\omega)$ would give us three sources of possible randomness. Here we will show how to explicitly compute the stochastic gradient with these three sources of randomness.

\paragraph{Stochastic Functional Gradients.}
Naturally, to iteratively update $f$ in a stochastic manner, we need to sample instances from the labeled dataset as well as the whole distribution. Different from DSG, we here randomly sample a pair of data points, from the labeled and the unlabeled data distributions, respectively. Then we can obtain stochastic functional gradients for $R(f)$ with these two data points as follow,
\begin{align}
g(\cdot) = f(\cdot) + \xi(\cdot)
\end{align}
where $\xi(\cdot)$ is the gradient contributed by the loss from both labeled and unlabeled data. It can be computed using the chain rule
\begin{align}
\xi(\cdot) =Cl'(f(x^l),y^l)k(x^l, \cdot)+C^{*}u'(f(x^u))k(x^u, \cdot)
\end{align}
where $x^l$, $x^u$ are sampled from the labeled dataset and unlabeled distribution $P(x)$ respectively. Next we will plugging the random feature approximation technique described in the previous section.

\paragraph{Random Feature Approximation.}
According to Eq. (\ref{random features}), when we use stochastically generated random feature $\omega$, we can further approximate $\xi(\cdot)$ as:
\begin{align}
&\xi(\cdot) \approx \zeta(\cdot)
\\
=& Cl'(f(x^l),y^l)\phi_{\omega}(x^l)\phi_{\omega}(\cdot)
+C^{*}u'(f(x^u))\phi_{\omega}(x^u)\phi_{\omega}(\cdot)\nonumber
\end{align}
note that $\xi(\cdot) = \mathbb{E}_{\omega}[\zeta(\cdot)]$. This leads to an unbiased estimator of the original functional gradient with three layers of stochasticity, \emph{i.e.},
\begin{align}
\nabla R(f) = \mathbb{E}_{(x^l,y^l)\in \mathcal{L}} \mathbb{E}_{x^u\sim P(x)}\mathbb{E}_{\omega}(\zeta(\cdot)) + f(\cdot)
\end{align}
Since three random events occur per iteration, \emph{i.e.} $x^l$, $x^u$, $\omega$, we call our approximate functional gradient as triply stochastic functional gradient.

\paragraph{Update Rules.}
In the $t$-th iteration, the triply stochastic (functional) gradient update rule for $f$ is:
\begin{align}
\label{f_upd}
f_{t+1}(\cdot) = f_t(\cdot) - \gamma_t (\zeta_t(\cdot)+f_t(\cdot))  =\sum_{i=1}^{t}a^i_t \zeta_i(\cdot)
\end{align}
where $\gamma$ denotes the step size and  the initial value $f_{1}(\cdot)=0$. It is straight forward to calculate that $a_t^i=-\gamma_i\prod_{k=i+1}^{t}(1-\gamma_k)$.
Ideally, if we could somehow compute the stochastic (functional) gradients $\xi_j(\cdot)$, the update rule becomes:
\begin{align}
\label{h_upd}
&h_{t+1}(\cdot) = h_t(\cdot) - \gamma_t (\xi_t(\cdot) + h_t(\cdot)) = \sum_{i=1}^{t}a^i_t \xi_i(\cdot)
\end{align}
where we have used $h_{t+1}$ instead of $f_{t+1}$ to distinguish from the triply stochastic (functional) gradient update rule and $h_1(\cdot)=0$. However, to avoid the expense of kernel computation, our algorithm will use the triply stochastic update rule Eq. (\ref{f_upd}) instead of Eq. (\ref{h_upd}).

\subsection{Algorithm}
Based on the above triply stochastic gradient update rules (\ref{f_upd}), we provide the TSGS$^3$VM training and prediction procedures  in Algorithms \ref{algo.1} and \ref{algo.2} receptively.
Notice that directly computing all the random features still needs
a large amount of memory.
Following the pseudo-random number generators setting of \cite{dai2014scalable}, our random feature generator is initialized by a predefined seed according to iteration. Thus, TSGS$^3$VM does not need to save the random feature matrix which makes it more memory friendly.
In the $i$-th iteration, our method will execute the following steps.

\begin{enumerate}
	\item Sample  data pair (lines 2-3 in Algorithm \ref{algo.1}): Stochastically sample a labeled sample $(x^l_i,y^l_i)$ and an unlabeled sample $x^u_i$ from different distribution respectively.
	\item
	Sample random features (line 4 in Algorithm \ref{algo.1}): Stochastically sample $\omega^i \sim \mathbb{P}(\omega)$ with seed $i$ and generate random features.
	\item
	Update coefficients (lines 5-8 in Algorithm \ref{algo.1}): Evaluate function value and update $f$ according to Eq. (\ref{f_upd}).
\end{enumerate}

\begin{remark}
	For each iteration, TSGS$^3$VM needs $O(mT)$ operations to evaluate function value, since evaluating the function value needs generating $m$ random features ($O(m)$) for $T$ times. Thus, the total computational complexity of TSGS$^3$VM is $O(mT^2)$. Due to the use of random features and pseudo-random method, TSGS$^3$VM only requires $O(T)$ memory, where $T$ is the iteration number.
\end{remark}

\begin{algorithm}[tb]
	\renewcommand{\algorithmicrequire}{\textbf{Input:}}
	\renewcommand{\algorithmicensure}{\textbf{Output:}}
	\caption{\textbf{TSGS${^3}$VM} \textbf{Train}}
	\label{algo.1}
	\begin{algorithmic}[1]
		\REQUIRE${\cal L}, P(x), \mathbb{P}(\omega), \phi_\omega(x), u(f(x)), C, C^{*}$
		\FOR {$i=1,\cdots, T$}
		\STATE	Sample $(x_i^l, y_i^l) \sim \mathcal{L}$
		\STATE	Sample $x_i^u \sim P(x)$
		\STATE	Sample $\omega_i \sim \mathbb{P}(\omega)$ with seed $i$
		\STATE	$f(x^l_i)=\textbf{Predict}(x^l_i, \{\alpha_j\}_{j=1}^{i-1})$
		\STATE	$f(x^u_i)=\textbf{Predict}(x^u_i, \{\alpha_j\}_{j=1}^{i-1})$
		\STATE	$\alpha_i=-\gamma_i (Cl'(f(x^l_i), y^l_i)\phi_{\omega_i}(x^l_i)+C^{*}u'(f(x^u_i))\phi_{\omega_i}(x^u_i))$
		\STATE	$\alpha_{j}=(1-\gamma_i)\alpha_{j}$, for $j=1,\cdots,i-1$
		\ENDFOR
		\ENSURE $\{\alpha_i\}_{i=1}^T$.
	\end{algorithmic}
\end{algorithm}

\begin{algorithm}[tb]
	\renewcommand{\algorithmicrequire}{\textbf{Input:}}
	\renewcommand{\algorithmicensure}{\textbf{Output:}}
	\caption{\textbf{TSGS${^3}$VM} \textbf{Predict}}
	\label{algo.2}
	\begin{algorithmic}[1]
		\REQUIRE {$\mathbb{P}(\omega), \phi_\omega(x), x, \{\alpha_i\}_{i=1}^{T}$}
		\STATE Set $f(x)=0$
		\FOR {$i=1,\cdots, T$}
		\STATE	Sample $\omega_i \sim \mathbb{P}(\omega)$ with seed $i $
		\STATE	$f(x)=f(x)+\alpha_i \phi_{\omega_i}(x)$
		\ENDFOR
		\ENSURE {$f(x)$}
	\end{algorithmic}
\end{algorithm}

\section{Theoretical Guarantees}
\begin{figure}[!htbp]
	\centering
	\includegraphics[scale=0.38]{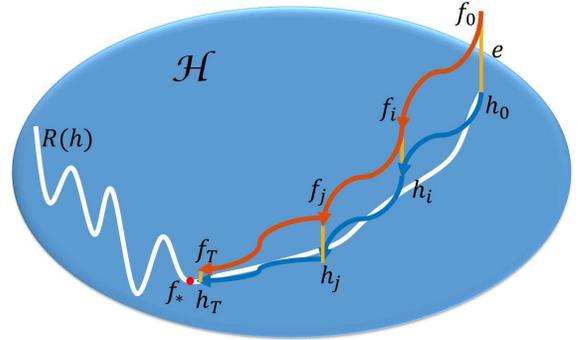}
	\caption{Illustration of how TSGS$^3$VM converge to a stationary point, where $e$ denotes for the error between $f_t$ and $h_t$, the white line denote the objective value $R(h)$. In this toy model we assume all horizontal points in $\cal H$ have the same objective value.}
	
	\label{fig:TSG}
\end{figure}
We follow the common goal of non-convex analysis \cite{ghadimi2013stochastic,gu2018faster,huo2018training} to bound $\mathbb{E}||\nabla R(f)||^2$, which means that the objective function will converge (in expectation) to a stationary point $f_{*}$.
When we use the hypothetical update rule (\ref{h_upd}), $h_t$ will always be inside of $\cal H$. However, because we could only use random features to approximate $h_t$ with $f_t$, we face the risk that functional $f_t$ could be outside of $\cal H$.
As a consequence, $\mathbb{E}||\nabla R(f)||^2_{\cal H}=0$ is not the stationary point of the objective function (\ref{S3VM2}). From Eq. (\ref{f_upd}) and Eq. (\ref{h_upd}), it is obvious that every update of $h_t$ happens implicitly with an update of $f_t(x)$. According to this relationship, we proposed to divide the analysis in two parts. As illustrated in Fig. \ref{fig:TSG}, for a general non-convex optimization problem $R(h)$,  we prove that the $h_{t+1}$  converges to a stationary point $f_*$ (\emph{i.e.}, $\mathbb{E}||\nabla R(h_{t+1})||^2_{\cal H}<\epsilon_1$) firstly. Then we prove that $f_{t+1}(x)$  keeps close to its hypothetic twin $h_{t+1}(x)$ for any $x \in \cal X$ (\emph{i.e.}, $|f_{t+1}(x)-h_{t+1}(x)|^2<\epsilon_2$).
\begin{table}[htbp]
	\centering
	\begin{tabular}{ccccc}
		\hline
		Dataset & Dimensionality & Samples  & Source\\
		\hline
		CodRNA &       8        &  59,535   &{LIBSVM}\\
		W6a&  300& 49749&{LIBSVM}\\
		IJCNN1  &       22       &  49,990   &{LIBSVM}\\
		SUSY   &      18      &  5,000,000   &{LIBSVM} \\
		Skin   &      3      &  245,057   &{LIBSVM} \\
		Higgs&  28& 1,100,000&{LIBSVM}\\
		\hline
		Dota2&  16& 102,944&{UCI}\\
		HEPMASS  & 28& 10,500,000&{UCI}\\
		\hline
	\end{tabular}
	\caption{Datasets used in the experiments.}
	\label{tab:dataset}
\end{table}
\begin{figure*}[htb]
	
	\centering
	\begin{subfigure}[b]{0.24\textwidth}
		\includegraphics[width=1.6in]{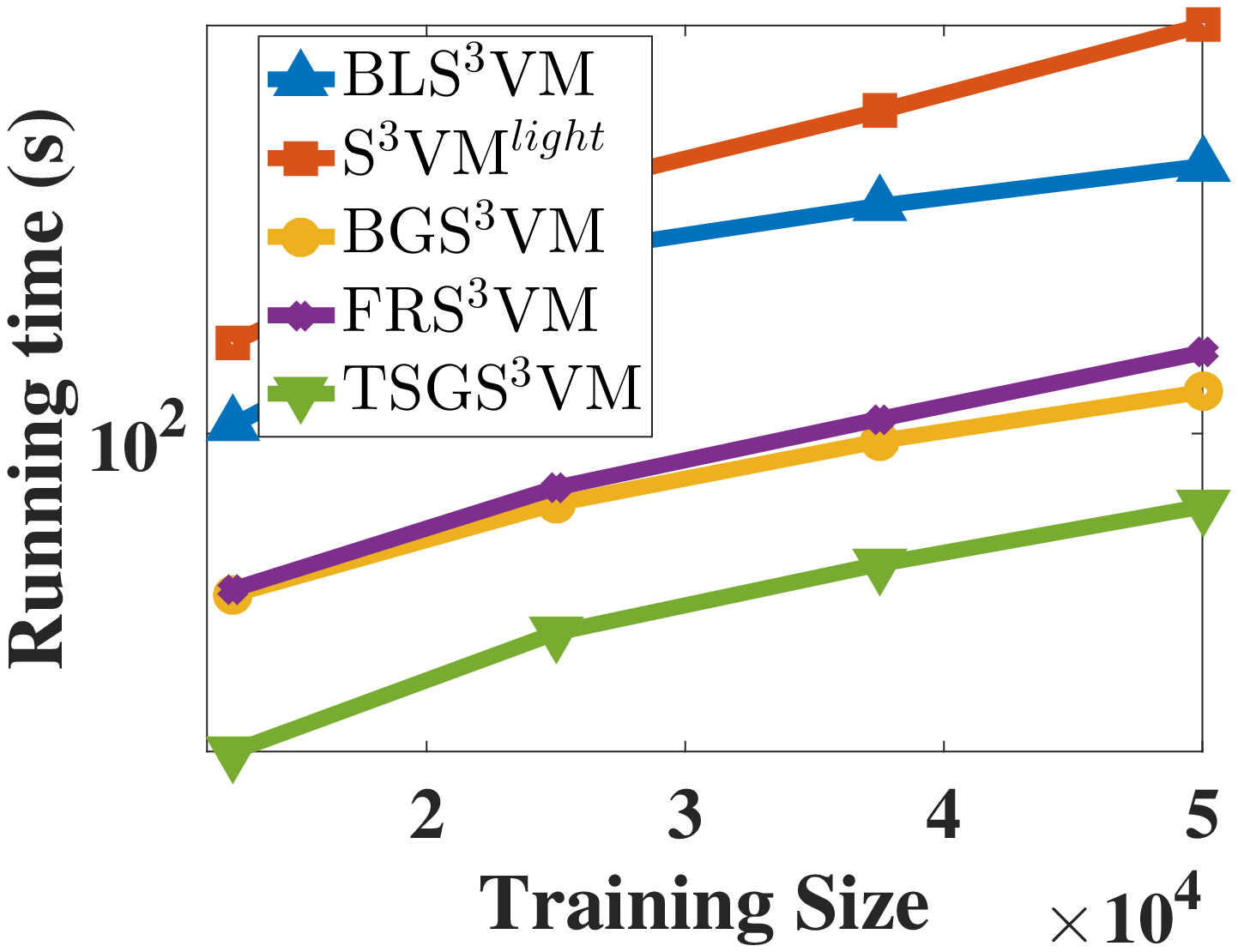}
		\caption{CodRNA}
	\end{subfigure}
	\begin{subfigure}[b]{0.24\textwidth}
		\includegraphics[width=1.6in]{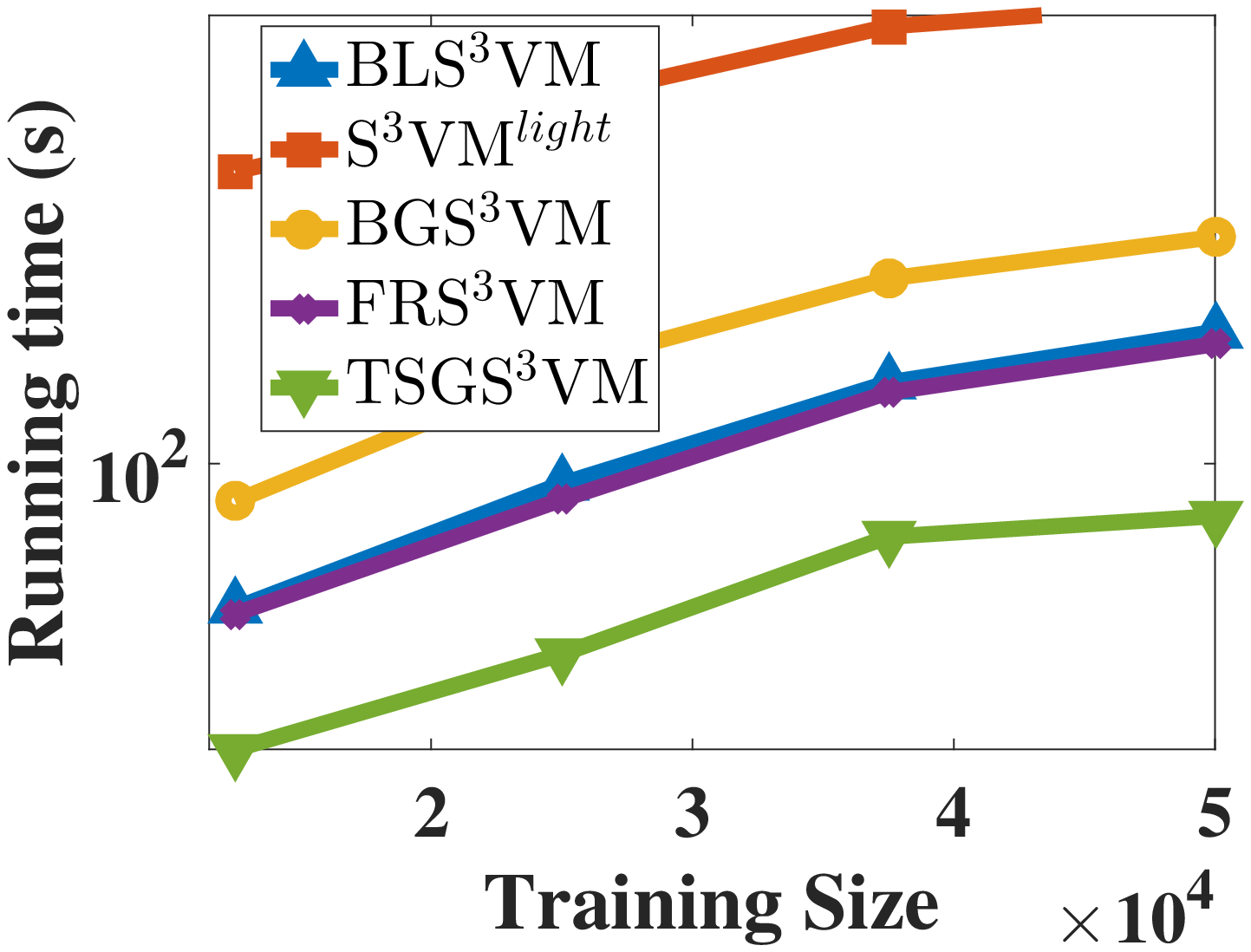}
		\caption{W6a}
	\end{subfigure}
	\begin{subfigure}[b]{0.24\textwidth}
		\includegraphics[width=1.6in]{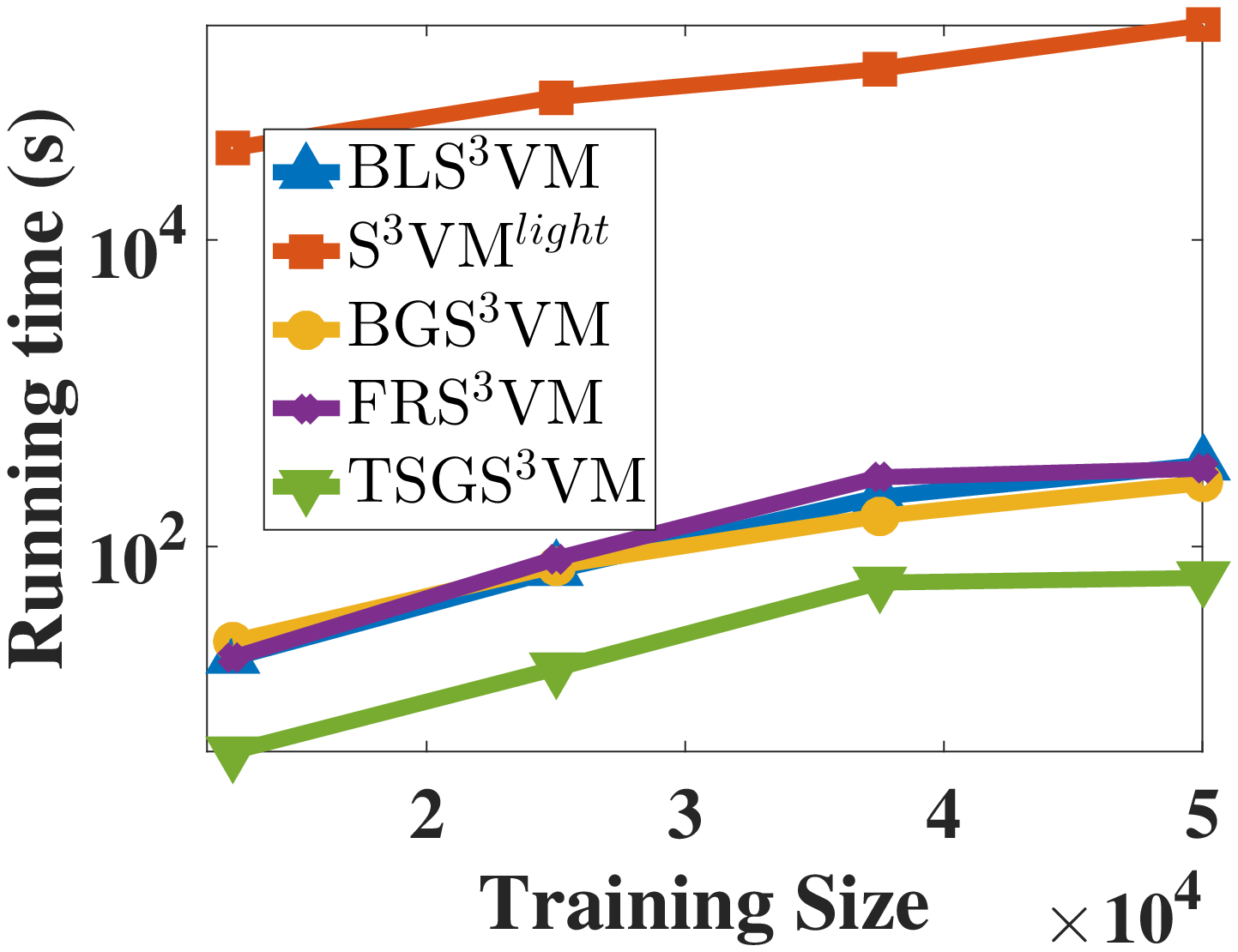}
		\caption{IJCNN1}
	\end{subfigure}
	\begin{subfigure}[b]{0.24\textwidth}\label{fig:susy}
		\includegraphics[width=1.6in]{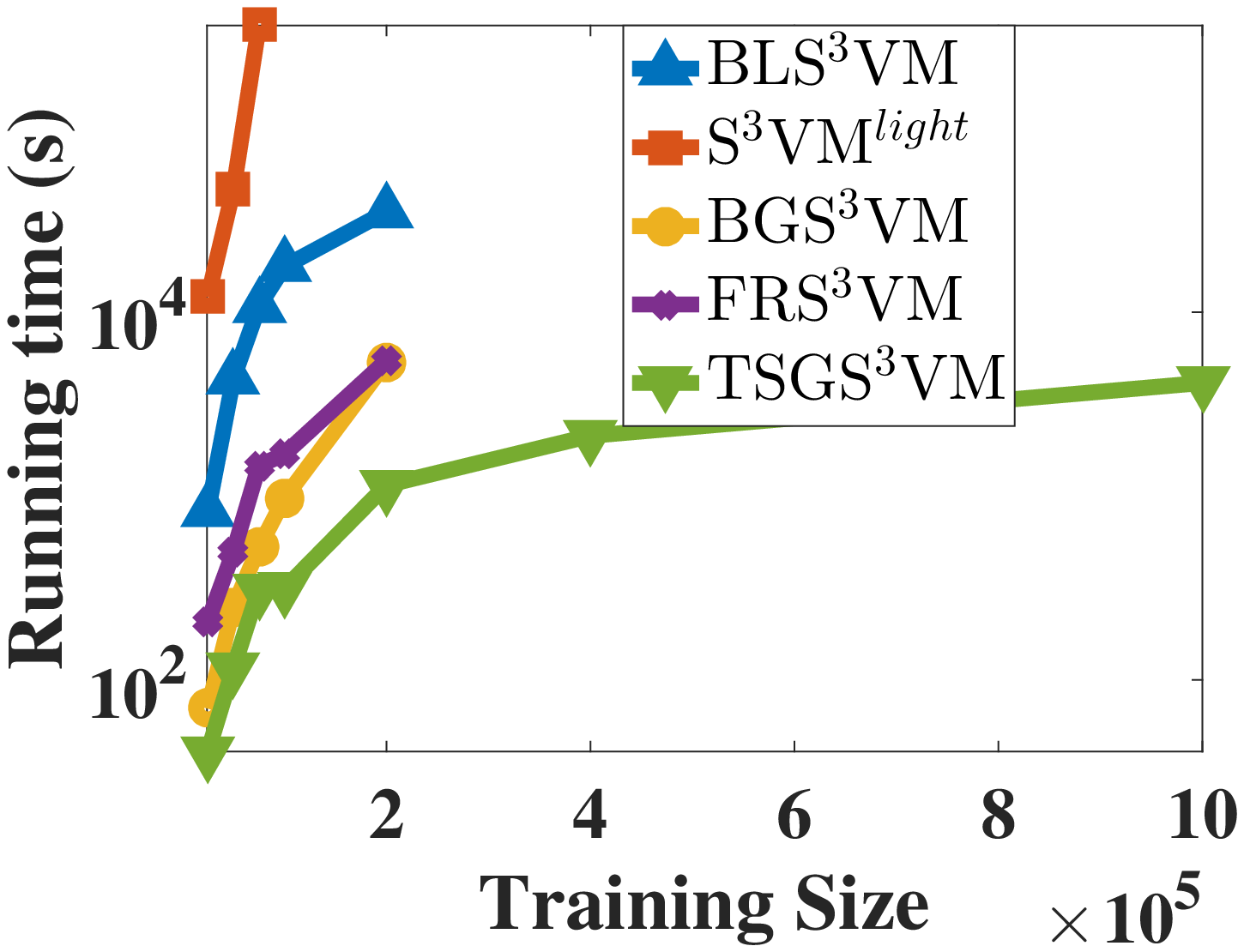}
		\caption{SUSY}
	\end{subfigure}
	\begin{subfigure}[b]{0.24\textwidth}
		\includegraphics[width=1.6in]{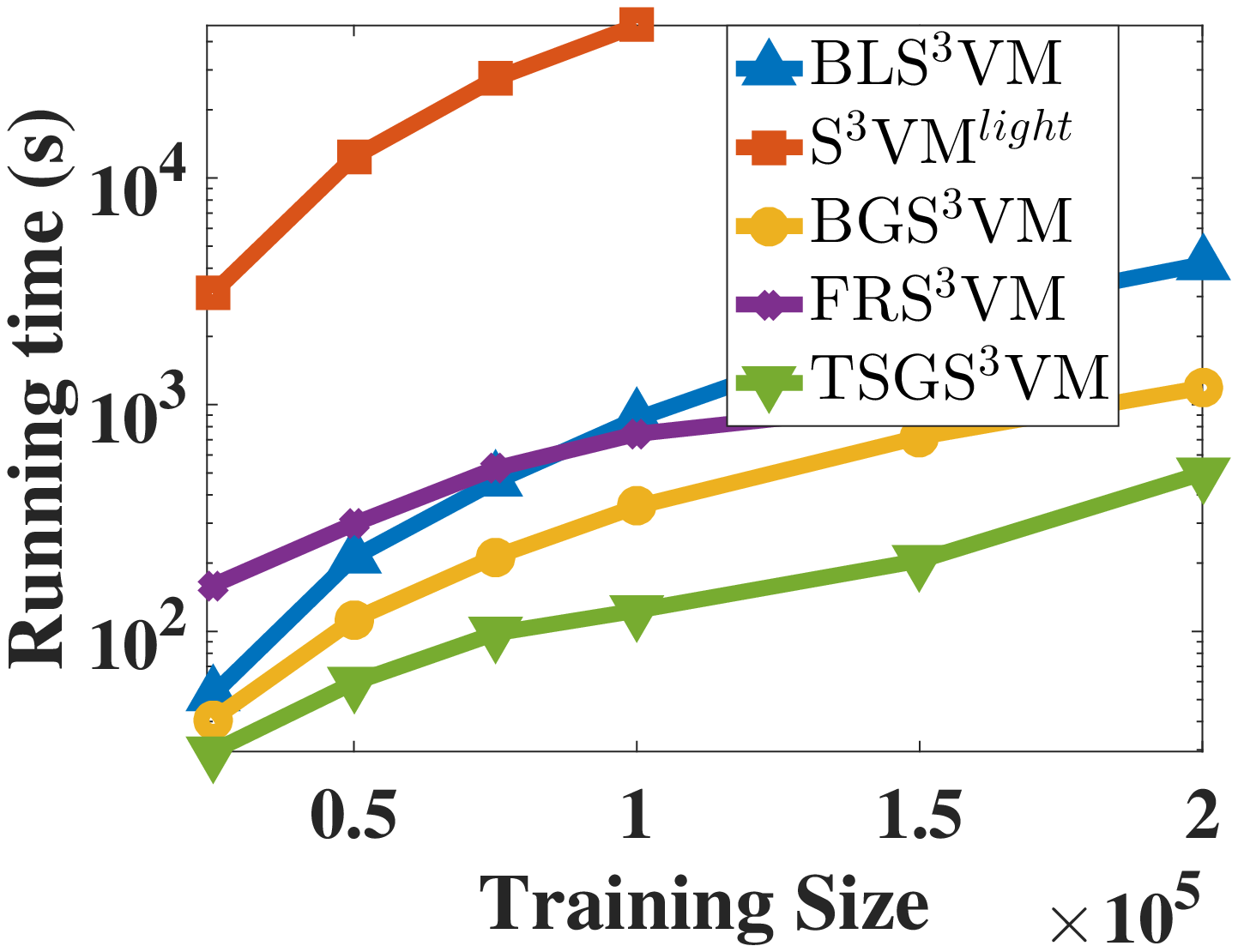}
		\caption{Skin}
	\end{subfigure}
	\begin{subfigure}[b]{0.24\textwidth}
		\includegraphics[width=1.6in]{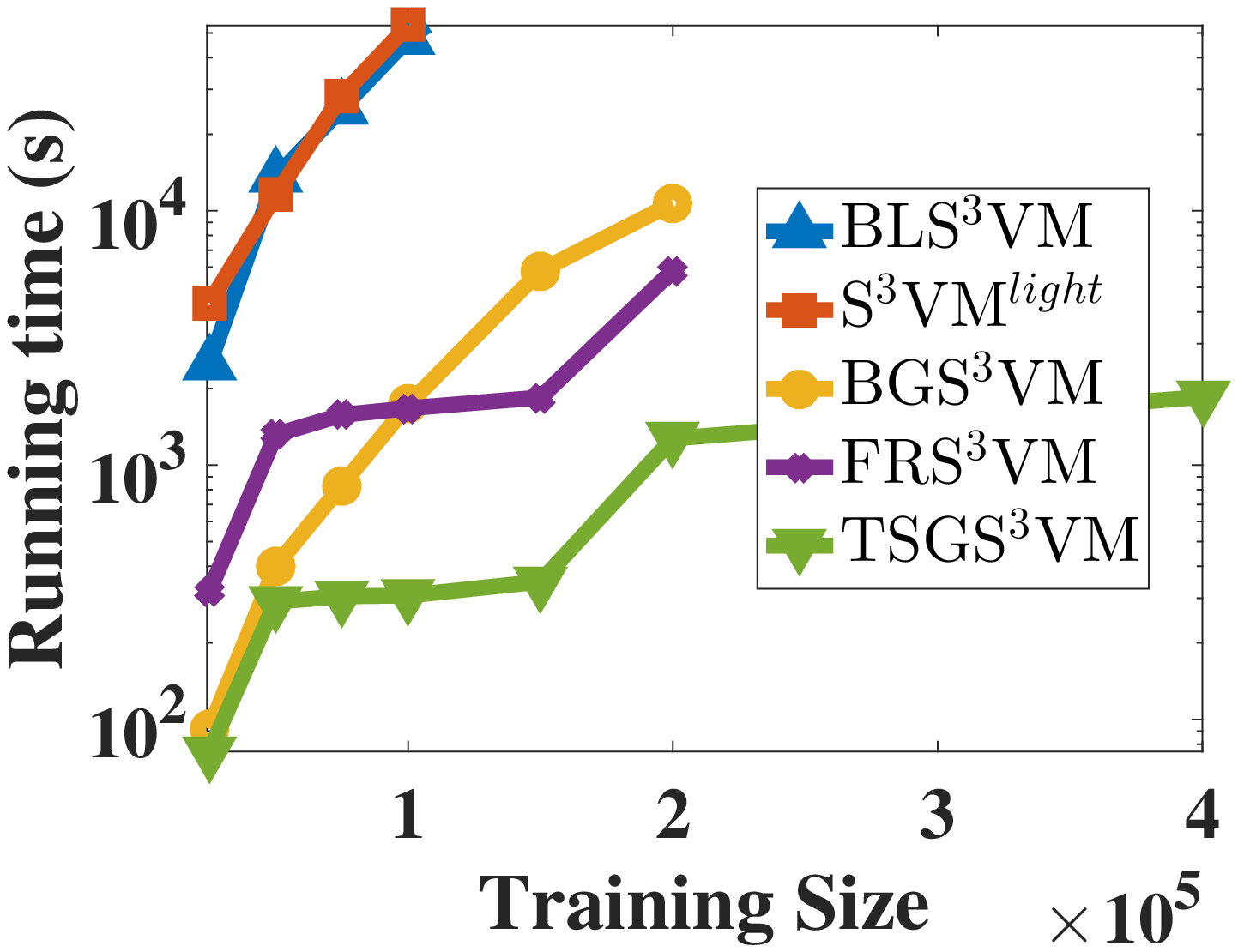}
		\caption{Higgs}
	\end{subfigure}
	\begin{subfigure}[b]{0.24\textwidth}
		\includegraphics[width=1.6in]{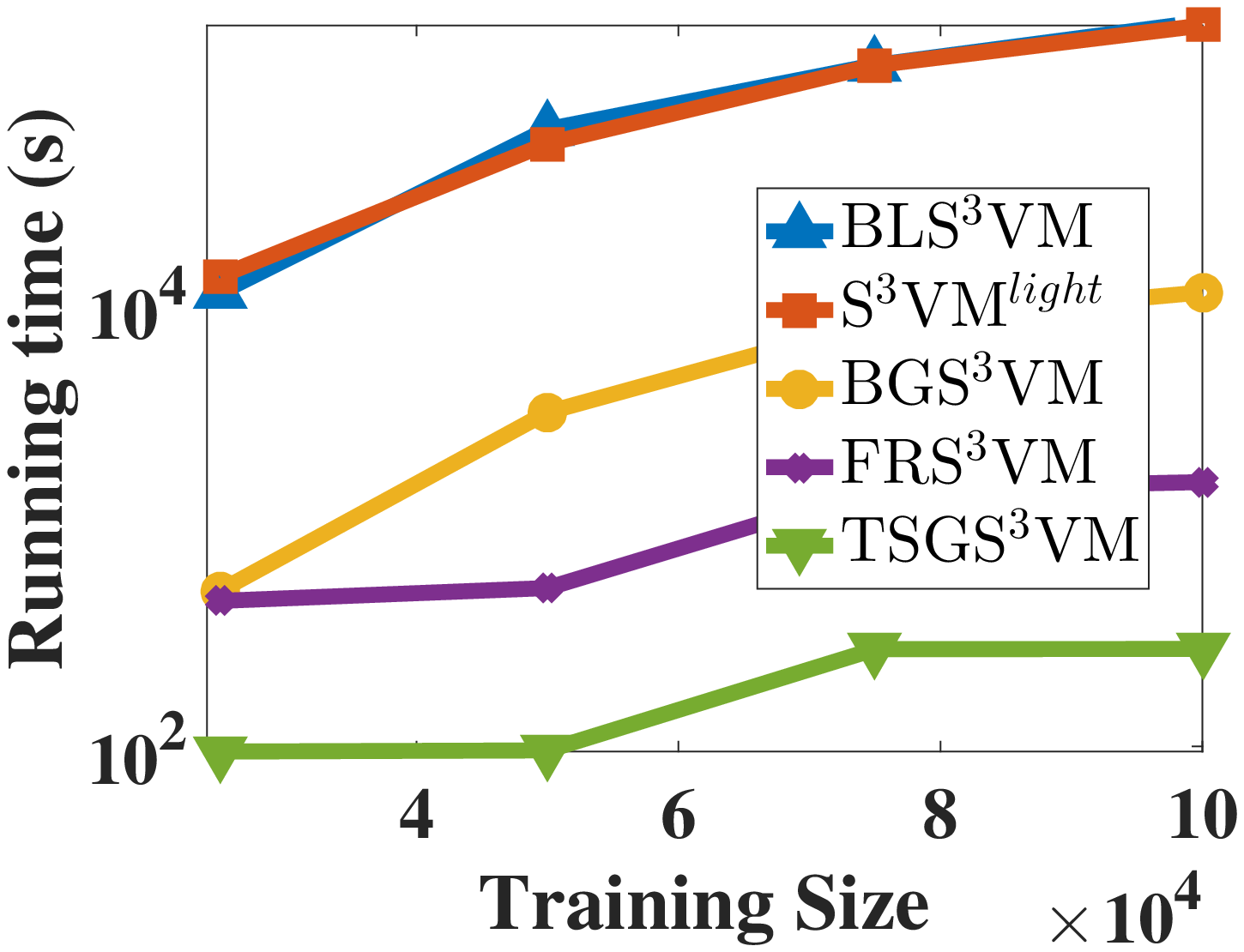}
		\caption{Dota2}
	\end{subfigure}
	\begin{subfigure}[b]{0.24\textwidth}\label{fig:hepmass}
		\includegraphics[width=1.6in]{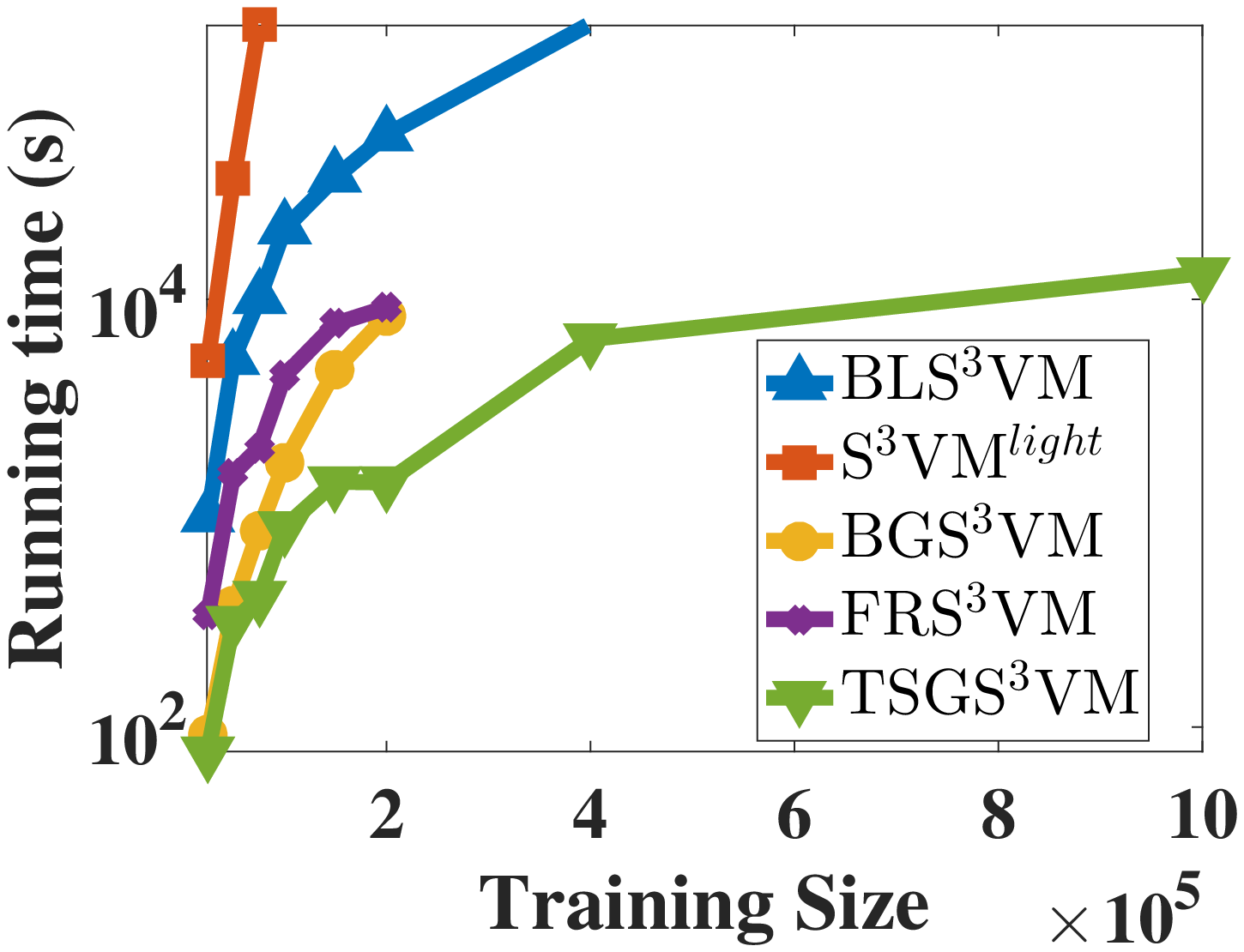}
		\caption{HEPMASS}
	\end{subfigure}
	
	\caption{Running time of different S$^3$VM solvers v.s. training size  on the eight  benchmark data sets, where the lines of BLS$^3$VM and S$^3$VM$^{light}$ are incomplete on several datasets due to the corresponding implementations crash on the large-scale training set.
	}
	\label{fig:time-size}
\end{figure*}

\begin{figure*}[!ht]
	\centering
	\includegraphics[width = 0.8\linewidth]{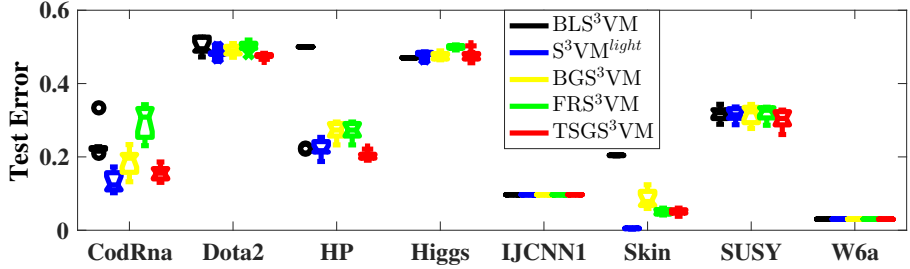}
	\caption{The boxplot of test error for different methods on different datasets.}
	\label{fig:test}
\end{figure*}

Our analysis is built upon the following assumptions which are standard for the analysis of non-convex optimization and DSG \cite{dai2014scalable}.
\begin{assumption}
	(Lipschitzian gradient) The gradient function $\nabla R(f)$  is Lipschitzian, that is to say
	\begin{align}
	||\nabla R(f) -\nabla R(g)||_{\cal H} \le L||f-g||_{\cal H}, \forall f,g \in \cal H
	\end{align}
\end{assumption}
\begin{assumption}
	(Lipschitz continuity) $l(r,v)$ is $L'$\textbf{-Lipschitz continuous} in terms of its 1st argument. $u(r,v)$ is $U'$\textbf{-Lipschitz continuous} in terms of its 1st argument. We further denote $M'=CL'+C^{*}U'$.
\end{assumption}
\begin{assumption}\label{assumption3}
	(Bound of derivative) The derivatives are bounded: $|l'|<M^l$ and $|u'|<M^u$, where $l'$ and $u'$ is the derivative of $l(r,v)$ and $u(r)$ w.r.t. the 1st argument respectively. We further denote $M=CM^l+C^{*}M^u$.
\end{assumption}

\begin{assumption}\label{assumption4}
	(Bound of kernel and random features) We have an upper bound for the kernel value, $k(x,x') \le \kappa$. There is an upper bound of random feature norm, \emph{i.e.}, $|\phi_{\omega}(x)\phi_{\omega}(x')| \le \phi$.
\end{assumption}
Suppose the total number of iterations is $T$, we introduce our main theorems as below. All the detailed proofs are provided in our Appendix.
\begin{theorem}
	\label{theorem1}
	For any $x \in \cal X$, fix $\gamma = \frac{\theta}{T^{3/4}}$ with $0<\theta\le T^{3/4}$, we have
	\begin{equation}
	\mathbb{E}_{x^l,x^u, \omega_t}\Big[\Big|f_{t}(x) - h_{t}(x)\Big|^2 \Big] \le \dfrac{D}{T^{1/2}}
	\end{equation}
	where $D = \theta^2M^2(\sqrt{\kappa}+\sqrt{\phi})^2$.
\end{theorem}
\begin{remark}
	The error between $f_{t+1}$ and $h_{t+1}$ is mainly induced by random features.
	Theorem \ref{theorem1} shows that this error has the convergence rate of $O(1/\sqrt{T})$ with proper step size.
\end{remark}

\begin{theorem}
	\label{theorem2}
	For fixed $\gamma=\frac{\theta}{T^{3/4}}$, $0<\theta\le T^{3/4}$, we have that
	\begin{align}
	\mathbb{E}_{x^l,x^u, \omega_t}[||\nabla R(h_t)||^2_{\cal H}]\le \frac{E}{T^{1/4}}+\frac{F}{{T}^{3/4}}
	\end{align}
	where $E=\frac{1}{\theta}[R(h_1)-R^*]+\theta M^2M'(\sqrt{\kappa}+\sqrt{\phi})\kappa, F=2\theta M^2L\kappa$,  $R^*$ denotes the optimal value of (\ref{S3VM2}).
\end{theorem}
\begin{remark}
	Instead of using the convexity assumption  in \cite{dai2014scalable}, Theorem \ref{theorem2} uses Lipschitzian gradient assumption to build the relationship between gradients and the updating functions $h_{t+1}$. Thus, we can bound each error term of  $\mathbb{E}_{x^l,x^u, \omega_t}[||\nabla R(h_t)||^2_{\cal H}]$ as shown in  Appendix. Note that compared to the strong assumption (\emph{i.e.}, the good initialization)  used in \cite{xie2015scale}, the assumptions used in our proofs are  weaker and more realistic.
\end{remark}

\section{Experiments and Analysis}
In this section, we will evaluate the practical performance of TSGS$^3$VM when comparing against other state-of-the-art solvers.

\subsection{Experimental Setup}
To show the advantage our TSGS$^3$VM for large-scale S$^3$VM learning, we conduct the experiments on large scale datasets to compare TSGS$^3$VM with other state-of-the-art algorithms in terms of predictive accuracy and time consumption. Specifically, the compared algorithms in our experiments are summarized as follows\footnote{BLS$^3$VM and S$^3$VM$^{light}$ can be found in http://pages.cs.wisc.edu/~jerryzhu/ssl/software.html}.
\begin{enumerate}
	\setlength{\itemsep}{-1pt}
	\item   \emph{\textbf{BLS$^3$VM}} \cite{collobert2006large}: The state-of-art S$^3$VM algorithm based on CCCP  and SMO algorithm \cite{cai2012generalized}.	
	\item  \textit{\textbf{S$^3$VM$^{light}$}} \cite{joachims1999transductive}: The implementation in the popular S$^3$VM$^{light}$ software. It is based on the local combinatorial search guided by a label switching procedure.
	\item  \emph{\textbf{BGS$^3$VM}}\cite{le2016budgeted}: Our implementation of BGS$^3$VM in MATLAB.
	\item  \emph{\textbf{FRS$^3$VM}}: Standard SGD with fixed random features.
	\item  \emph{\textbf{TSGS$^3$VM}}: Our proposed S$^3$VM algorithm via triply stochastic gradients.
\end{enumerate}

\paragraph{Implementation.}
We implemented the TSGS$^3$VM algorithm in MATLAB. For the sake of efficiency, our TSGS$^3$VM implementation also uses a mini-batch setting. We perform experiments on Intel Xeon E5-2696 machine with 48GB RAM.
The Gaussian RBF kernel $k(x,x')=\exp(-\sigma||x-x'||^2)$ and the loss function $u=\max\{0,1-|r|\}$ was used for all algorithms.
5-fold cross-validation was used to determine the optimal settings (test error) of the model parameters (the regularization factor C and the Gaussian kernel parameter $\sigma$), the parameters $C^{*}$ was set to $C\frac{n^l}{n^u}$.  Specifically, the unlabeled dataset was divided evenly to 5 subsets, where one of the subsets and all the labeled data are used for training, while the other 4 subsets are used for testing.
Parameter search was done on a 7$\times$7 coarse grid linearly spaced in the region $\{\log_{10} C , \log_{10}  \sigma)|  -3 \le \log_{10} C \le 3, -3 \le \log_{10} \sigma \le 3\}$ for all methods.
For TSGS$^3$VM, the step size $\gamma$ equals $\frac{1}{\eta}$, where $0 \le \log_{10}{\eta} \le 3$ is searched after $C$ and $\sigma$. Besides, the number of random features is set to be $\lceil\sqrt{n}\rceil$ and the batch size is set to 256.
The test error was obtained by using these optimal model parameters for all the methods. To achieve a comparable accuracy to our TSGS$^3$VM, we set the minimum budget sizes $B_l$ and $B_u$ as $100$ and $0.2*n_u$ respectively for BG$S^3$VM.
We stop TSGS$^3$VM and BG$S^3$VM after one pass over the entire dataset. We stop FRS$^3$VM after 10 pass over the entire dataset to achieve a comparable accuracy.
All results are  the average of 10 trials.

\paragraph{Datasets.}
Table \ref{tab:dataset} summarizes the {8} datasets used in our experiments. They are
from LIBSVM\footnote{https://www.csie.ntu.edu.tw/~cjlin/libsvmtools/datasets/} and
UCI\footnote{http://archive.ics.uci.edu/ml/datasets.html} repositories.
Since all these datasets are originally labeled, we intentionally randomly sample 200 labeled instances and treat the rest of data as unlabeled to make a semi-supervised learning setting.





\subsection{Experimental Results}
Fig. \ref{fig:time-size} shows the test error v.s. the training size for different algorithms. 
The results clearly show that TSGS$^3$VM runs much faster than other methods. Specifically, Figs. \ref{fig:time-size}d and \ref{fig:time-size}h confirm the high efficiency of TSGS$^3$VM even on the datasets with one million samples. Besides, TSGS$^3$VM requires low memory benefiting  from pseudo-randomness for generating random features, while BLS$^3$VM and S$^3$VM$^{light}$ would be often out of memory on  large scale datasets.

Fig. \ref{fig:test} shows the test error of different methods.  The results were obtained at the optimal hyper-parameters for different algorithms.
From the figure, it is clear that TSGS$^3$VM achieves similar generalization performance as that of BLS$^3$VM, S$^3$VM$^{light}$, and BGS$^3$VM methods which confirm that TSGS$^3$VM converge well in practice. Besides, TSGS$^3$VM achieves better generalization performance than FRS$^3$VM, because TSGS$^3$VM has the advantage that it would automatically use more and more random features (for each data $x$) as the number of iterations increases.

Based on these results, we conclude that TSGS$^3$VM is much more efficient and scalable than these algorithms while retaining the similar generalization performance.

\section{Conclusion}
In this paper, we provide a novel triply stochastic gradients algorithm for kernel S$^3$VM to make it scalable.
We establish new theoretic analysis for TSGS$^3$VM which guarantees that TSGS$^3$VM can efficiently converge to a stationary point for a general non-convex learning problem under weak assumptions. As far as we know, TSGS$^3$VM is the first  work that offers non-convex analysis for DSG-like algorithm without a strong initialization assumption.
Extensive experimental results on a variety of benchmark datasets demonstrate the superiority of our proposed TSGS$^3$VM.


\section*{Acknowledgments}
H.H. was partially supported by U.S. NSF IIS 1836945, IIS 1836938, DBI 1836866, IIS 1845666, IIS 1852606, IIS 1838627, IIS 1837956. B.G. was partially supported by the National Natural Science Foundation of China (No: 61573191), and the Natural Science Foundation  (No. BK20161534), Six talent peaks project (No. XYDXX-042) in Jiangsu Province.

\appendix
\section{Proof of Theorem \ref{theorem1}}\label{proof1}
To proof theorem \ref{theorem1}, we first give lemma \ref{lemma1} as follows
\begin{lemma}
	\label{lemma1}
	For any $x \in \cal X$, consider a fix step size $\gamma$, and $0<\gamma \le1$, we have that
	\begin{equation}
	\mathbb{E}_{x^l, x^u, \omega_t}\Big[\Big|f_{t}(x) - h_{t}(x)\Big|^2 \Big] \le B^2_{1,t+1}
	\end{equation}
	where $B^2_{1,t+1} := M^2(\sqrt{\kappa}+\sqrt{\phi})^2t\gamma^2$, and $B_{1,1}=0$.
\end{lemma}
\textbf{Proof.} We denote $W_i(x)=W_i(x;x_i^l,x_i^u,\omega_i)=a_t^i(\zeta^i(x)-\xi^i(x))$. Based on the definition and assumptions in main body, $W_i(x)$ is bounded as follows
\begin{align}
&\quad\quad|W_i(x)|
\nonumber\\&\le\quad c_i= \quad|a_t^i|(|\zeta_i(x)|+|\xi_i(x)|)
\nonumber\\&=\quad|a_t^i|\big[|Cl'(f(x_i^l),y_i^l)k(x_i^l,x)+C^{*}u'(f(x_i^u))k(x_i^u,x)|\nonumber\\&\quad+|Cl'(f(x_i^l),y_i^l)\phi_{\omega_i}(x_i^l)\phi_{\omega_i}(x) \nonumber\\&\quad+C^{*}u'(f(x_i^u))\phi_{\omega_i}(x_i^l)\phi_{\omega_i}(x)|\big]
\nonumber\\&\le\quad |a_t^i| \big[(CM^l+C^{*}M^u)(\sqrt{\kappa}+\sqrt{\phi})\big]
\nonumber\\&=\quad M(\sqrt{\kappa}+\sqrt{\phi})|a_t^i|
\end{align}
Based on the definition of $a_t^t$, $|a_t^t|\le \gamma$, and for any $i$ we have $|a_t^i|\le \gamma \prod_{j=i+1}^{t}(1-\gamma)\le \gamma$ due to $0< \gamma \le1$. Consequently, $\sum_{i=1}^{t}|a_t^i|^2\le t\gamma^2$. Then we have $\mathbb{E}_{x^l_t,x^u_t, \omega_t}\Big[\Big|f_{t}(x) - h_{t}(x)\Big|^2 \Big] \le M^2(\sqrt{\kappa}+\sqrt{\phi})^2\sum_{i=1}^{t}|a_{t}^i|^2\le M^2(\sqrt{\kappa}+\sqrt{\phi})^2t\gamma^2$.
We obtain the Lemma \ref{lemma1}. Taking $\gamma = \frac{\theta}{T^{3/4}}$, we have 
\begin{align}
B^2_{1,t+1} &:= \quad M^2(\sqrt{\kappa}+\sqrt{\phi})^2t\frac{\theta^2}{T^{3/2}}\nonumber\\
&\le \quad M^2(\sqrt{\kappa}+\sqrt{\phi})^2T\frac{\theta^2}{T^{3/2}}\nonumber\\
&\le \quad M^2(\sqrt{\kappa}+\sqrt{\phi})^2\frac{\theta^2}{T^{1/2}}
\end{align}
then we obtain theorem \ref{theorem1}.

\section{Proof of Theorem \ref{theorem2}}\label{proof2}
\textbf{Proof} For the sake of simple notations, let us first denote the following three different gradient terms, which are
\begin{align}
g_t &=\quad \xi_t + h_{t} 
\nonumber\\&=\quad Cl'(f_t(x_t^l), y_t^l) k(x_t^l, \cdot) \nonumber\\
&\quad + C^{*}u'(f_t(x_t^u)) k(x_t^u, \cdot) +  h_t \nonumber\\
\hat{g}_t &=\quad \hat{\xi}_t + h_{t} 
\nonumber\\&=\quad Cl'(h_t(x_t^l), y_t^l) k(x_t^l, \cdot) \nonumber\\
&\quad + C^{*}u'(h_t(x_t^u)) k(x_t^u, \cdot) + h_t \nonumber\\
\nabla R(h_t) &=\quad \mathbb{E}_{x^l,x^u}[\hat{g}_t]
\nonumber\\&=\quad\mathbb{E}_{x^l,x^u}[Cl'(h_t(x_t^l), y_t^l) k(x_t^l, \cdot) \nonumber\\
&\quad + C^{*}u'(h_t(x_t^u)) k(x_t^u, \cdot)] \nonumber+  h_t\nonumber
\end{align}
Besides, we use $\mathbb{E}$ to denotes $\mathbb{E}_{x^l_t,x^u_t,\omega_t}$ for simplicity sometimes.
Similar as \cite{ghadimi2013stochastic,reddi2016stochastic}, using the assumption 1 and  $h_{t+1}=h_t - \gamma g_t, \forall t\ge 1$, for $t = 1,\cdots, T$ we have

\begin{align}
&\quad\quad R(h_{t+1})
\nonumber\\&\le \quad R(h_{t})+\langle \nabla R(h_t), h_{t+1}-h_{t}\rangle+\frac{L}{2}||h_{t+1}-h_t||^2_{\cal H}\nonumber\\
& =\quad R(h_{t})-\gamma\langle \nabla R(h_t),  g_t\rangle+\frac{L\gamma^2}{2}||g_t||^2_{\cal H} \nonumber\\
& =\quad R(h_{t})-\gamma\langle \nabla R(h_t),  g_t-\hat{g}_t+\hat{g}_t-\nabla R(h_t)
\nonumber\\&\quad+\nabla R(h_t)\rangle
+\frac{L\gamma^2}{2}||g_t||^2_{\cal H} \nonumber\\
& =\quad R(h_{t})-\gamma||\nabla R(h_t)||^2_{\cal H}+\gamma\langle \nabla R(h_t),  \hat{g}_t-g_t\rangle
\nonumber\\&\quad+\gamma\langle\nabla R(h_t),\nabla R(h_t)-\hat{g}_t\rangle+\frac{L\gamma^2}{2}||g_t||^2_{\cal H}
\end{align}


Summing up the above inequalities from and re-arranging the terms, we obtain
\begin{align}
&\quad\quad\frac{1}{T}\sum_{t=1}^{T}\gamma||\nabla R(h_t)||^2_{\cal H} 
\nonumber\\&\le\quad\frac{1}{T}\big[ R(h_1)-R(h_{T+1})+\sum_{t=1}^{T}\gamma\langle \nabla R(h_t),  \hat{g}_t-g_t\rangle 
\nonumber\\&\quad+ \sum_{t=1}^{T}\gamma\langle\nabla R(h_t),\nabla R(h_t)-\hat{g}_t\rangle+\sum_{t=1}^{T}\frac{L\gamma^2}{2}||g_t||^2_{\cal H}\big]
\nonumber\\&\le\quad \frac{1}{T}\big[R(h_1)-R^*+\sum_{t=1}^{T}\gamma\langle \nabla R(h_t),  \hat{g}_t-g_t\rangle 
\nonumber\\&\quad+ \sum_{t=1}^{T}\gamma\langle\nabla R(h_t),\nabla R(h_t)-\hat{g}_t\rangle+\sum_{t=1}^{T}\frac{L\gamma^2}{2}||g_t||^2_{\cal H}\big]
\end{align}
where $R^*$ is the optimal value of $R(h_t)$ and the last inequality follows from the fact that $R(h_{T+1})\ge R^*$. 
Taking expectations on both sides we have
\begin{align}
&\quad\quad\mathbb{E}[||\nabla R(h_t)||^2_{\cal H}]
\nonumber\\&\le\quad \frac{1}{T\gamma}\big[R(h_1)-R^*\big]+\mathbb{E}\big[\langle \nabla R(h_t),  \hat{g}_t-g_t\rangle \big]
\nonumber\\&\quad+ \mathbb{E}\big[\langle\nabla R(h_t),\nabla R(h_t)-\hat{g}_t\rangle\big]+\frac{L\gamma}{2}\mathbb{E}\big[||g_t||^2_{\cal H}\big]
\end{align}
Let us denote $\mathcal{H}_t=\sqrt{||\nabla R(h_t)||^2_{\cal H}}, \mathcal{M}_t=||g_t||^2_{\cal H},\mathcal{N}_t=\langle\nabla R(h_t),\nabla R(h_t)-\hat{g}_t\rangle$,$\mathcal{R}_t=\langle \nabla R(h_t),  \hat{g}_t-g_t\rangle$. 
From lemma \ref{lemma2}, we have the bound as follows 
\begin{align}\label{bound1}
\mathbb{E}[\mathcal{H}_t]^2&\le 4 \kappa M^2\\
\label{bound2}
\mathbb{E}[\mathcal{N}_t]&=0\\
\label{bound3}
\mathcal{M}_t&\le 4 \kappa M^2\\
\label{bound4}
\mathbb{E}[\mathcal{R}_t]&\le \kappa M M'B_{1,t}
\end{align}

Applying these bounds we give above leading to the refined recursion as follows 
\begin{align}
&\quad\quad\mathbb{E}[||\nabla R(h_t)||^2_{\cal H}]
\nonumber\\&\le\quad \frac{1}{T\gamma}\big[R(h_1)-R^*\big]+ \kappa M M'B_{1,t} +2L\gamma\kappa M^2\nonumber
\\&\le\quad \frac{1}{T\gamma}\big[R(h_1)-R^*\big]+ \kappa M^2M'\sqrt{t}\gamma (\sqrt{\kappa}+\sqrt{\phi})\nonumber
\\&\quad +2L\gamma\kappa M^2
\end{align}
Let $\gamma = \frac{\theta}{{T^{3/4}}}$, we have
\begin{align}
&\quad\quad\mathbb{E}[||\nabla R(h_t)||^2_{\cal H}]
\nonumber\\&\le\quad \frac{1}{T^{1/4}}\big[\frac{1}{\theta}[R(h_1)-R^*]\big]
+ \frac{1}{{T^{1/4}}} \theta\kappa  M^2M'(\sqrt{\kappa}+\sqrt{\phi})\nonumber
\\&\quad +\frac{1}{{T}^{3/4}}2\theta\kappa M^2L
\end{align}
Thus, we complete the proof.

\section{Lemma \ref{lemma2}}
\begin{lemma}
	\label{lemma2}
	In this lemma, we prove the inequalities (\ref{bound1})-(\ref{bound4})
\end{lemma}
\textbf{Proof} Similar to \cite{dai2014scalable}, given the definitions of $ \mathcal{M}_t,\mathcal{H}_t,\mathcal{N}_t,\mathcal{R}_t$ we have

(1) $\mathcal{M}_t\le 4 \kappa M^2$; 

This is because $\mathcal{M}_t=||g_t||^2_{\cal H}=||\xi_t+ h_t||_{\cal H}^2\le (||\xi_t||_{\cal H}+||h_t||_{\cal H})^2$.
We have 
\begin{align}\label{xi}
&\quad\quad||\xi_t||_{\cal H}
\nonumber\\&=\quad||Cl'(f_t(x_t^l),y_t^l)k(x_t^l,\cdot)+C^{*}u'(f_t(x_t^u))k(x_t^u,\cdot)||_{\cal H}
\nonumber\\&\le\quad||Cl'(f_t(x_t^l),y_t^l)k(x_t^l,\cdot)||_{\cal H}+||C^{*}u'(f_t(x_t^u))k(x_t^u,\cdot)||_{\cal H}
\nonumber\\&\le\quad CM^l\kappa^{1/2}+C^{*}M^u\kappa^{1/2} \le \kappa^{1/2}M
\end{align}
where the first inequality because triangle inequality, and from lemma \ref{lemma3} we have $||h_t||_{\cal H}\le {\kappa ^{1/2}M}$. Applying these bounds leads to $\mathcal{M}_t \le 4 \kappa M^2$

(2) $\mathbb{E}_{x^l_t,x^u_t,\omega_t}[\mathcal{H}_t]^2\le 4 \kappa M^2$; 
This is because 
\begin{align}
\mathbb{E}_{x^l_t,x^u_t,\omega_t}[\mathcal{H}_t]^2&=\quad||\mathbb{E}_{x^l_t,x^u_t,\omega_t}[\hat{\xi}_t]+\nu h_t||^2_{\cal H}
\nonumber\\&\le\quad (||\mathbb{E}_{x^l_t,x^u_t,\omega_t}[\hat{\xi}_t]||_{\cal H}+v||h_t||_{\cal H})^2\nonumber
\end{align}
where $||\mathbb{E}_{x^l_t,x^u_t,\omega_t}[\hat{\xi}_t]||_{\cal H} \le \kappa^{1/2}M$ according to (\ref{xi}), and from lemma \ref{lemma3} we have $||h_t||_{\cal H}\le {\kappa ^{1/2}M}$. Applying these bounds leads to $\mathbb{E}_{x^l_t,x^u_t,\omega_t}[\mathcal{H}_t]^2 \le 4 \kappa M^2$

(3) $\mathbb{E}_{x^l_t,x^u_t,\omega_t}[\mathcal{N}_t]=0$; This is because
\begin{align}
&\quad\quad\mathbb{E}_{x^l_t,x^u_t,\omega_t}[\mathcal{N}_t]
\nonumber\\&=\quad\mathbb{E}_{x^l_{t-1},x^u_{t-1},\omega_t}
[\mathbb{E}_{x^l_t,x^u_t}[\langle\nabla R(h_t),\nabla R(h_t)-\hat{g}_t\rangle|x^l_{t-1},x^u_{t-1},\omega_t]]
\nonumber\\&=\quad\mathbb{E}_{x^l_{t-1},x^u_{t-1},\omega_t}[\langle\nabla R(h_t),\mathbb{E}_{x^l_t,x^u_t}[\nabla R(h_t)-\hat{g}_t]\rangle]=0
\end{align}


(4) $\mathbb{E}_{x^l_t,x^u_t,\omega_t}[\mathcal{R}_t]= \le \kappa M M'B_{1,t}$; This is because
\begin{align}
&\quad\quad\mathbb{E}_{x^l_t,x^u_t,\omega_t}[\mathcal{R}_t]
\nonumber\\&=\quad\mathbb{E}_{x^l_t,x^u_t,\omega_t}[\langle \nabla R(h_t),  g_t-\hat{g}_t\rangle] \nonumber
\\&=\quad\mathbb{E}_{x^l_t,x^u_t,\omega_t}[\langle \nabla R(h_t),[Cl'(h_t(x_t^l),y_t^l)-Cl'(f_t(x_t^l),y_t^l)]
\nonumber\\&\quad\cdot k(x_t^l,\cdot)\rangle]
\nonumber\\&\quad+\mathbb{E}_{x^l_t,x^u_t,\omega_t}[\langle \nabla R(h_t),[C^{*}u'(h_t(x_t^u))-C^{*}u'(f_t(x_t^u))]
\nonumber\\&\quad\cdot k(x_t^u,\cdot)\rangle]
\nonumber\\& \le \quad\mathbb{E}_{x^l_t,x^u_t,\omega_t}[|Cl'(h_t(x_t^l),y_t^l)-Cl'(f_t(x_t^l),y_t^l)|
\nonumber\\&\quad\cdot||k(x_t^l,\cdot)||_{\cal H}\cdot||\nabla R(h_t)||] 
\nonumber\\&\quad+ \mathbb{E}_{x^l_t,x^u_t,\omega_t}[|C^{*}u'(h_t(x_t^u))-C^{*}u'(f_t(x_t^u))|
\nonumber\\&\quad\cdot||k(x_t^u,\cdot)||_{\cal H}\cdot||\nabla R(h_t)||] \nonumber
\end{align}
\begin{align}
&\le\quad C\kappa^{1/2}L'\cdot \mathbb{E}_{x^l_t,x^u_t,\omega_t}[|h_t(x_t^l)-f_t(x_t^l)|\cdot||\nabla R(h_t)||_{\cal H}] \nonumber
\nonumber\\&\quad+ C^{*}\kappa^{1/2}U'\cdot \mathbb{E}_{x^l_t,x^u_t,\omega_t}[|h_t(x_t^u)-f_t(x_t^u)|\cdot||\nabla R(h_t)||_{\cal H}] \nonumber\\&\le\quad C\kappa^{1/2}L'
\nonumber\\&\quad\cdot \sqrt{\mathbb{E}_{x^l_t,x^u_t,\omega_t}|h_t(x_t^l)-f_t(x_t^l)|^2}
\cdot\sqrt{\mathbb{E}_{x^l_t,x^u_t,\omega_t}||\nabla R(h_t)||_{\cal H}^2} 
\nonumber\\&\quad+ C^{*}\kappa^{1/2}U'
\nonumber\\&\quad\cdot \sqrt{\mathbb{E}_{x^l_t,x^u_t,\omega_t}|h_t(x_t^u)-f_t(x_t^u)|^2}
\cdot\sqrt{\mathbb{E}_{x^l_t,x^u_t,\omega_t}||\nabla R(h_t)||_{\cal H}^2} 
\nonumber\\&\le\quad C\kappa^{1/2}L'B_{1,t}\sqrt{\mathbb{E}_{x^l_t,x^u_t,\omega_t}[\mathcal{H}_t]^2}
\nonumber\\&\quad+C^{*}\kappa^{1/2}U'B_{1,t}\sqrt{\mathbb{E}_{x^l_t,x^u_t,\omega_t}[\mathcal{H}_t]^2}
\nonumber\\&\le\quad \kappa M (CL'+C^{*}U')B_{1,t}=\kappa M M'B_{1,t}
\end{align}
where the first and third inequalities are due to Cauchy-Schwarz Inequality, the second inequality is due to $L'$-Lipschitz continuity of $l'(\cdot,\cdot)$ and $U'$-Lipschitz continuity of $u'(\cdot,\cdot)$, the forth inequality is due to lemma \ref{lemma1} and fifth inequality is due to the bound of $\mathbb{E}_{x^l_t,x^u_t,\omega_t}[\mathcal{H}_t]^2$ we give above.

\section{Lemma \ref{lemma3}}
\begin{lemma}
	\label{lemma3}
	For any $t = 1\cdots T$, we have $||h_{t}||\le {\kappa^{1/2}M}$
\end{lemma}
\textbf{Proof} For $t=1$, according to the definition of $h_t$ we have $h_1=0, ||h_1||_{\cal H} = 0 \le {\kappa^{1/2}M}$.
For any $t=1\cdots T-1$ if we have $||h_{t}||_{\cal H}\le {\kappa^{1/2}M}$, then according to the definition of $h_t$ we have
\begin{align}
||h_{t+1}||_{\cal H}&=\quad||(1- \gamma)h_t-\gamma\xi_t(\cdot)||_{\cal H}\nonumber\\
&\le\quad (1- \gamma)||h_t||_{\cal H}+\gamma ||\xi_t(\cdot)||_{\cal H}\nonumber\\
&\le\quad (1- \gamma){\kappa^{1/2}M}+\gamma \kappa^{1/2}M\nonumber\\
&\le\quad {\kappa^{1/2}M}
\end{align}
The first inequality is because triangle inequality, the second inequality is because the assumption bound of $h_t$ and the bound of $\xi_t(\cdot)$ in (\ref{xi}).
Now we have lemma \ref{lemma3}.
\begin{figure*}[ht]
	\centering
	\begin{subfigure}[]{0.24\textwidth}
		\includegraphics[width=1.6in]{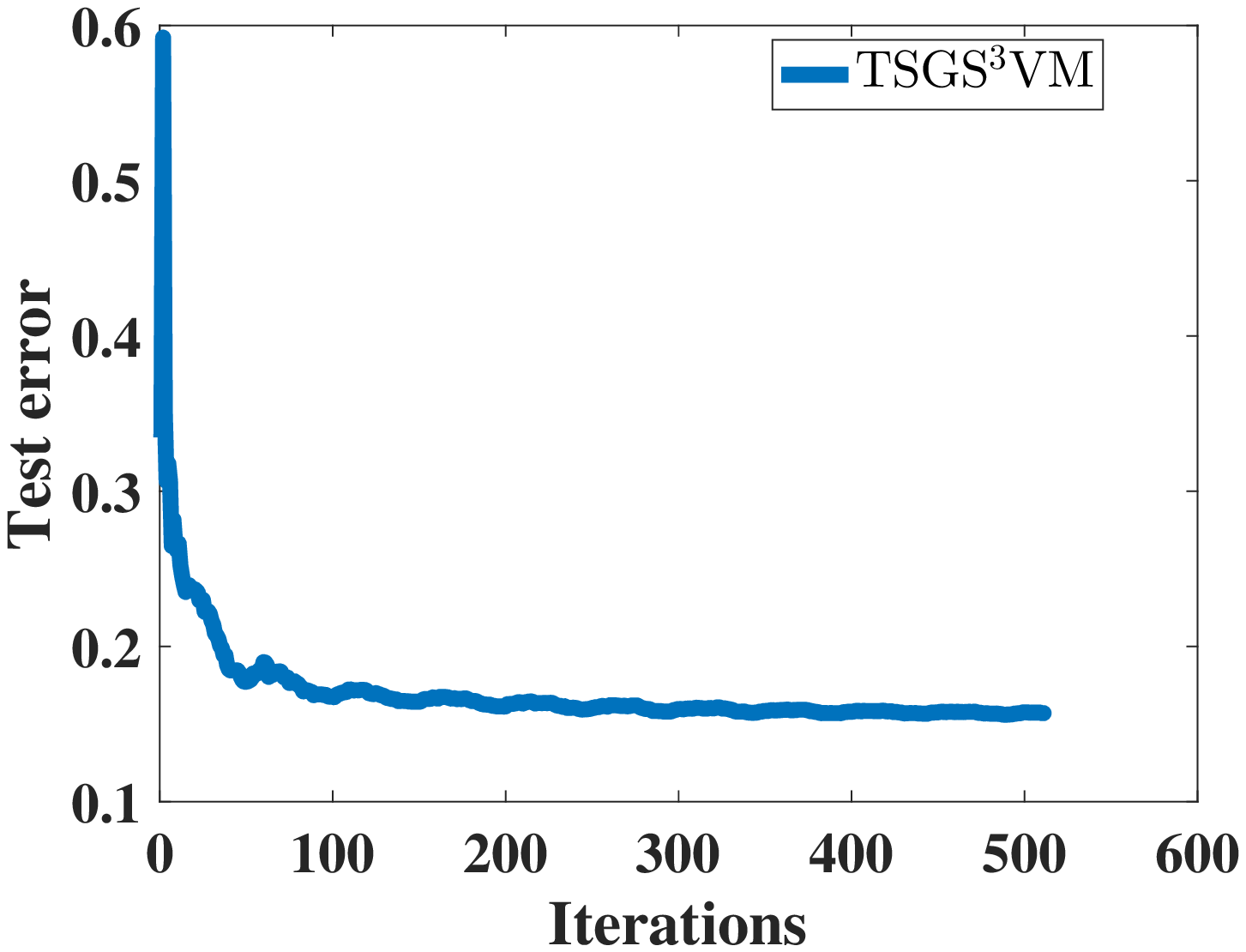}
		\caption{CodRNA}
	\end{subfigure}
	\begin{subfigure}[]{0.24\textwidth}
		\includegraphics[width=1.6in]{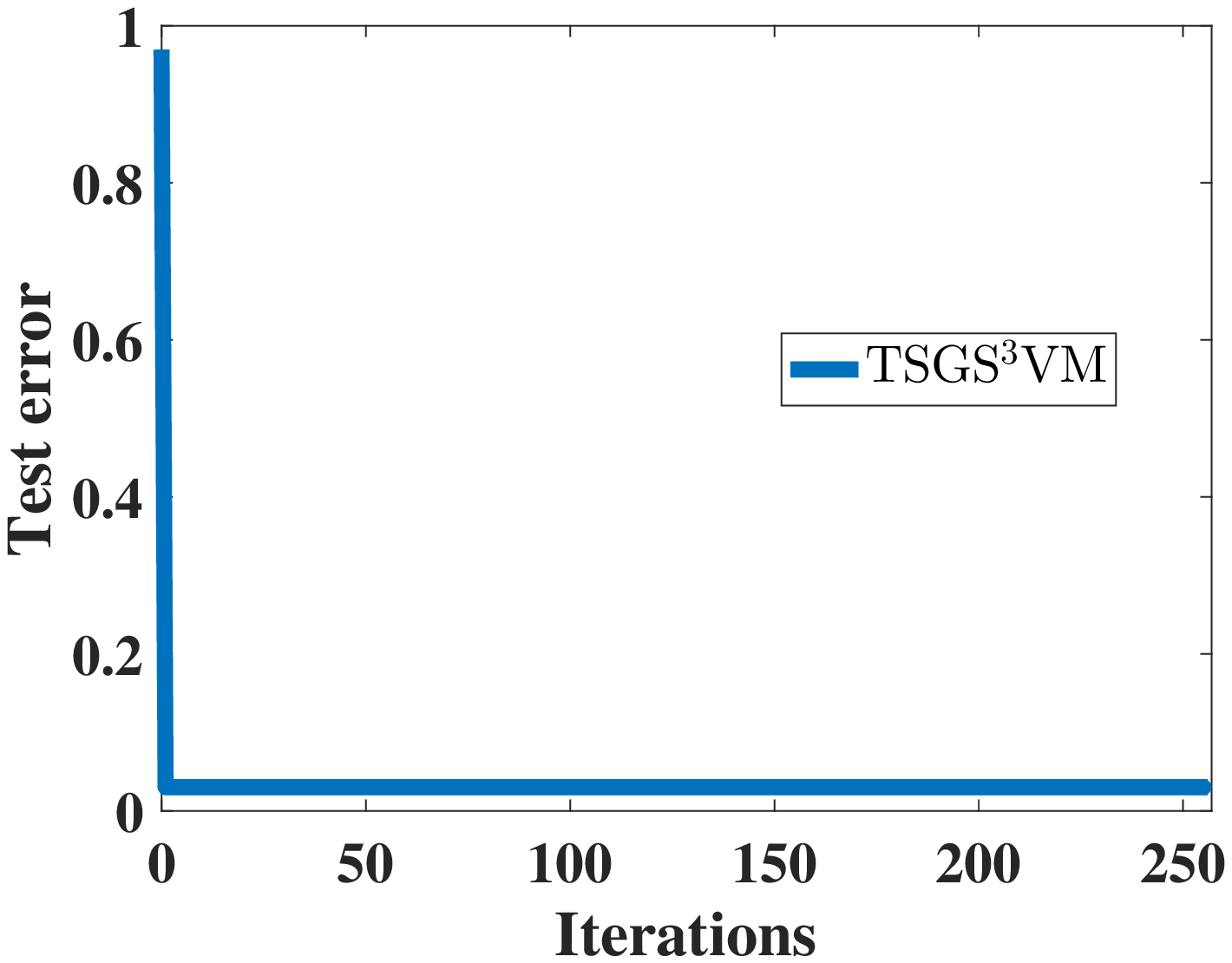}
		\caption{W6a}
	\end{subfigure}
	\begin{subfigure}[]{0.24\textwidth}
		\includegraphics[width=1.6in]{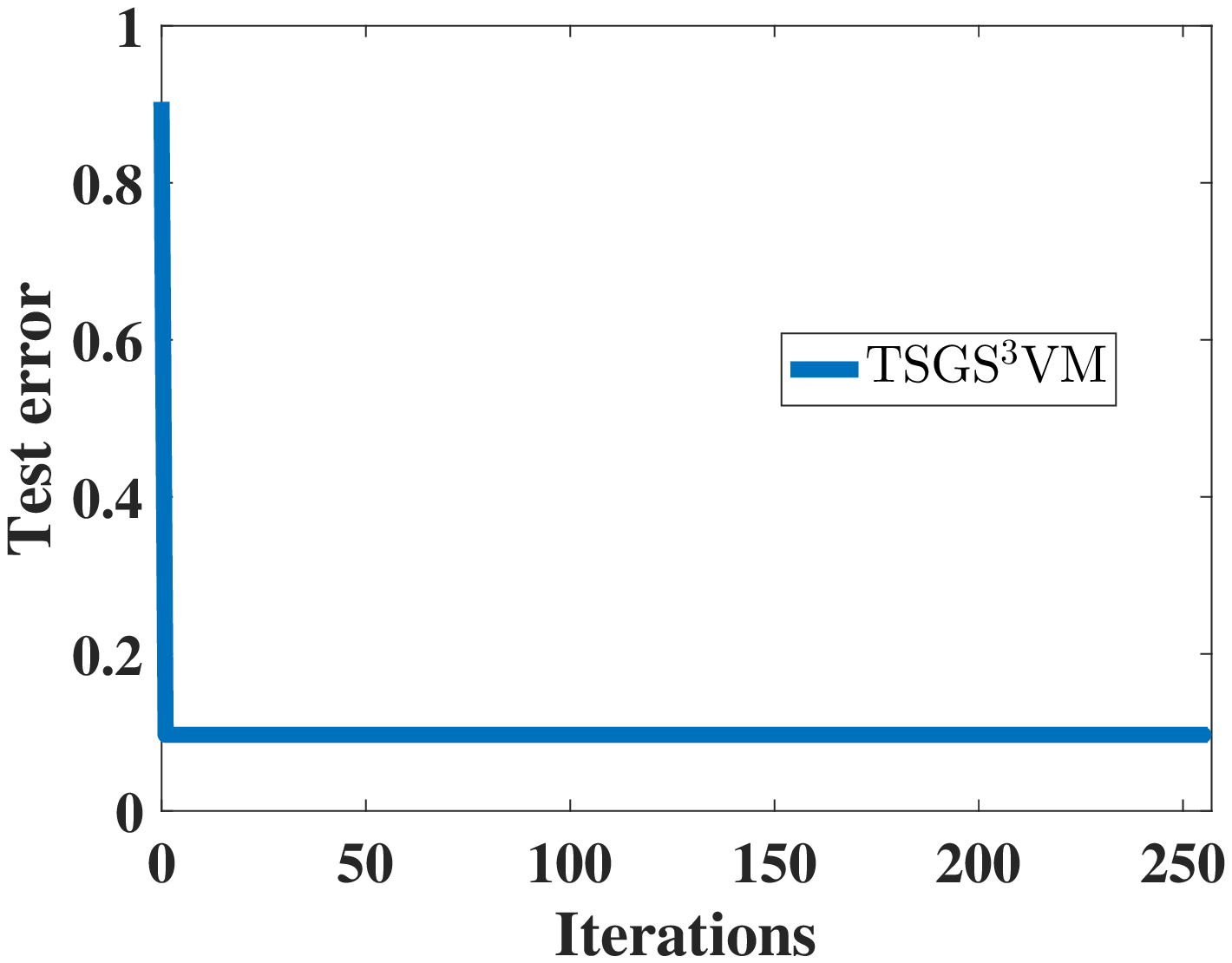}
		\caption{IJCNN1}
	\end{subfigure}
	\begin{subfigure}[]{0.24\textwidth}
		\includegraphics[width=1.6in]{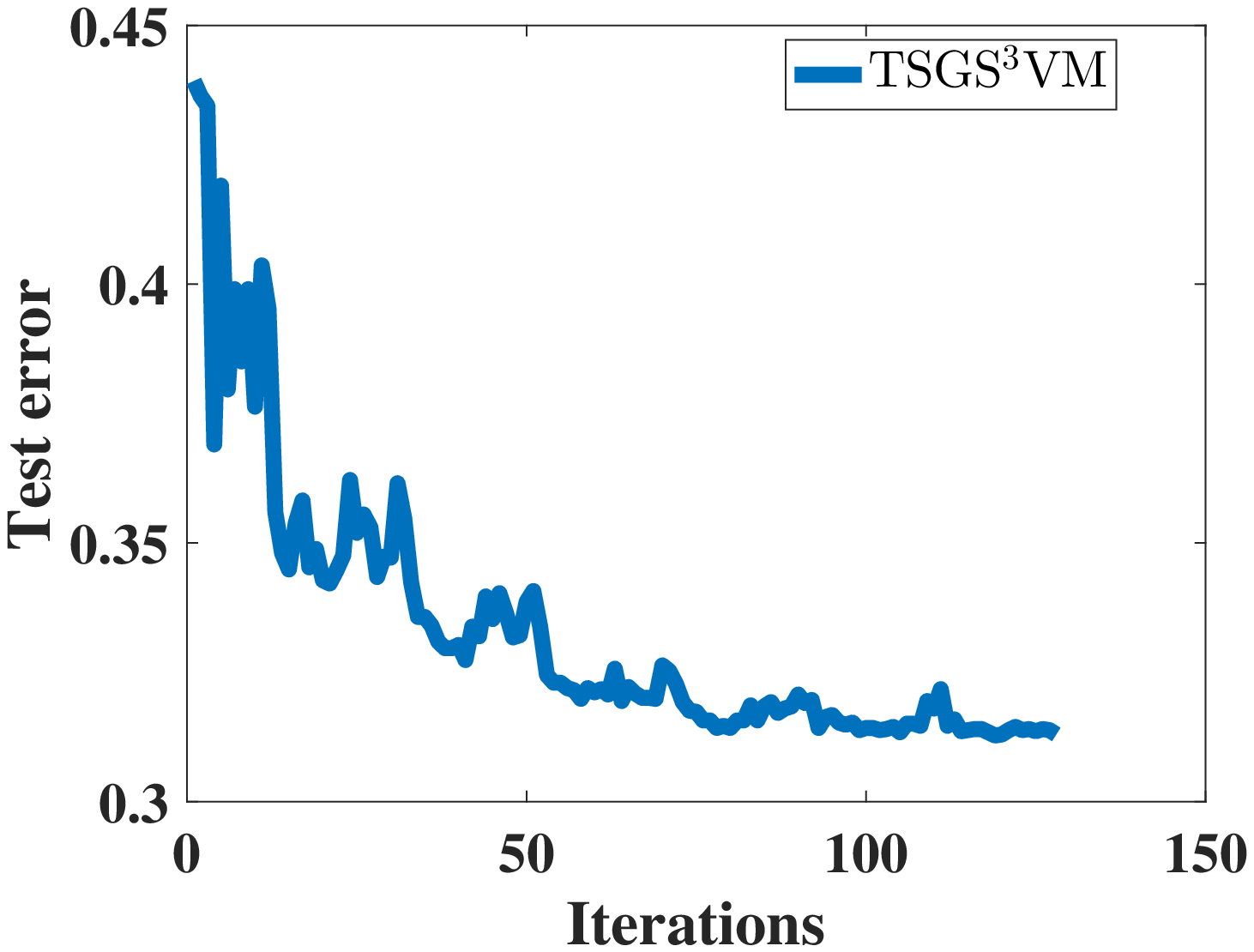}
		\caption{SUSY}
	\end{subfigure}
	\begin{subfigure}[]{0.24\textwidth}
		\includegraphics[width=1.6in]{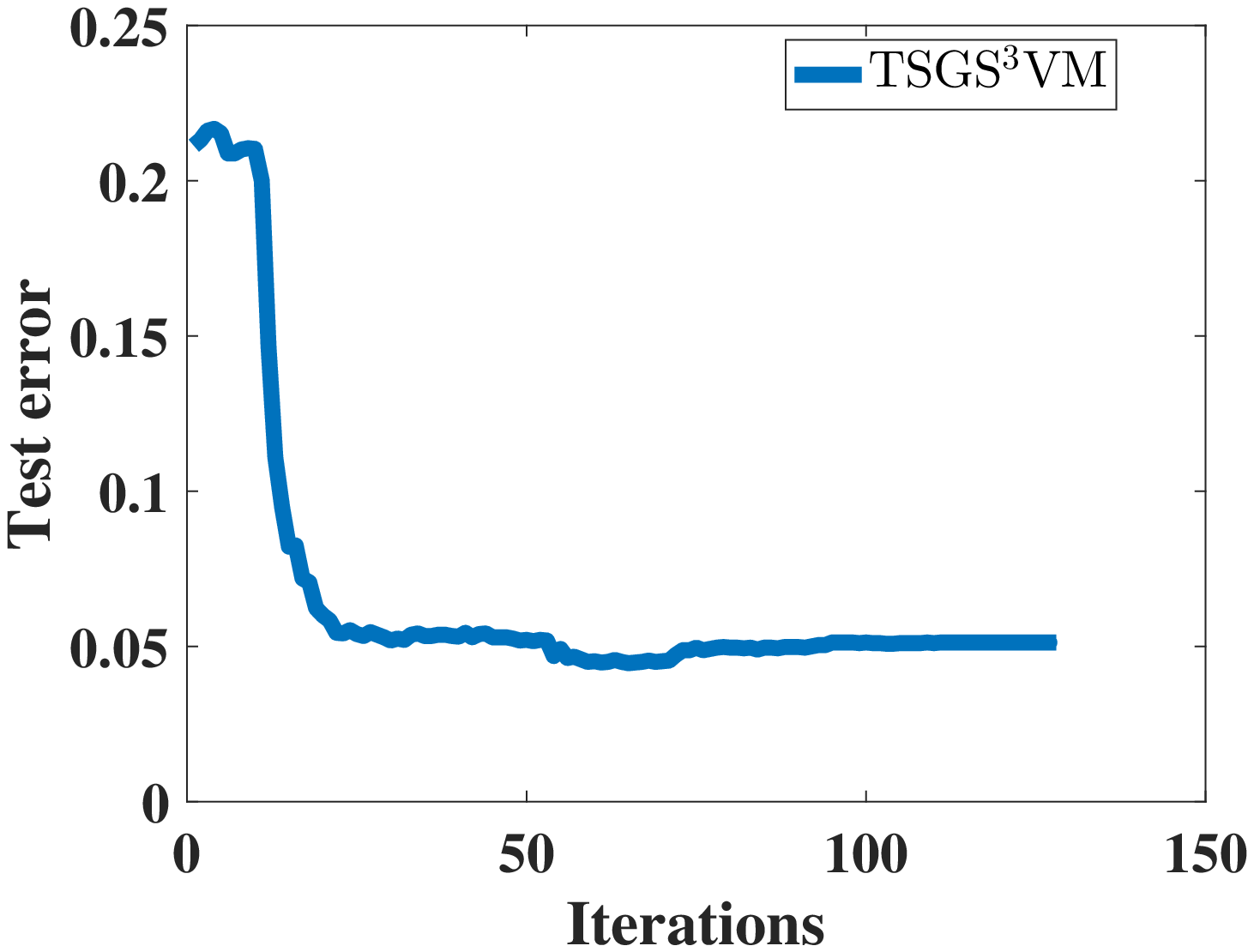}
		\caption{Skin}
	\end{subfigure}
	\begin{subfigure}[]{0.24\textwidth}
		\includegraphics[width=1.6in]{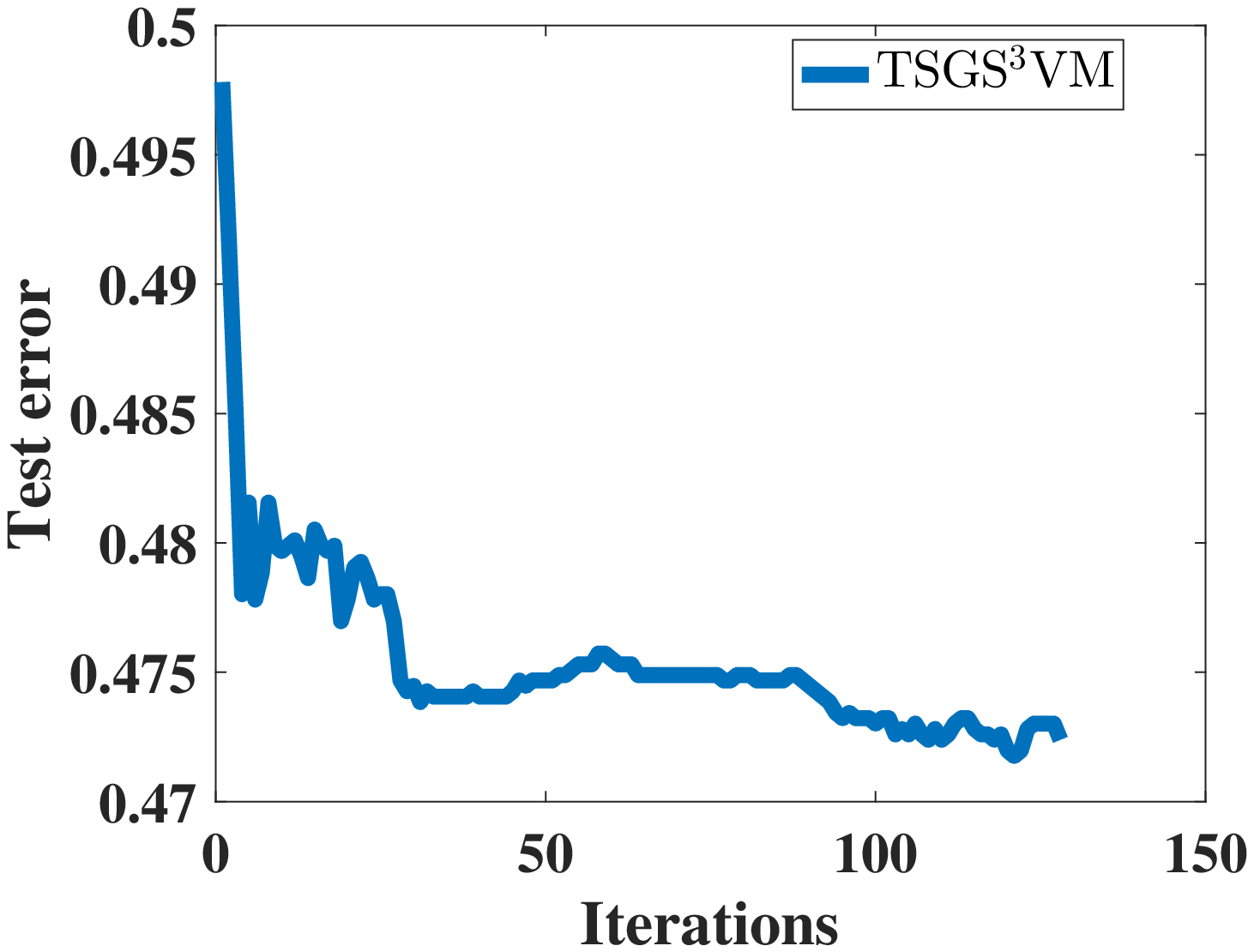}
		\caption{Higgs}
	\end{subfigure}
	\begin{subfigure}[]{0.24\textwidth}
		\includegraphics[width=1.6in]{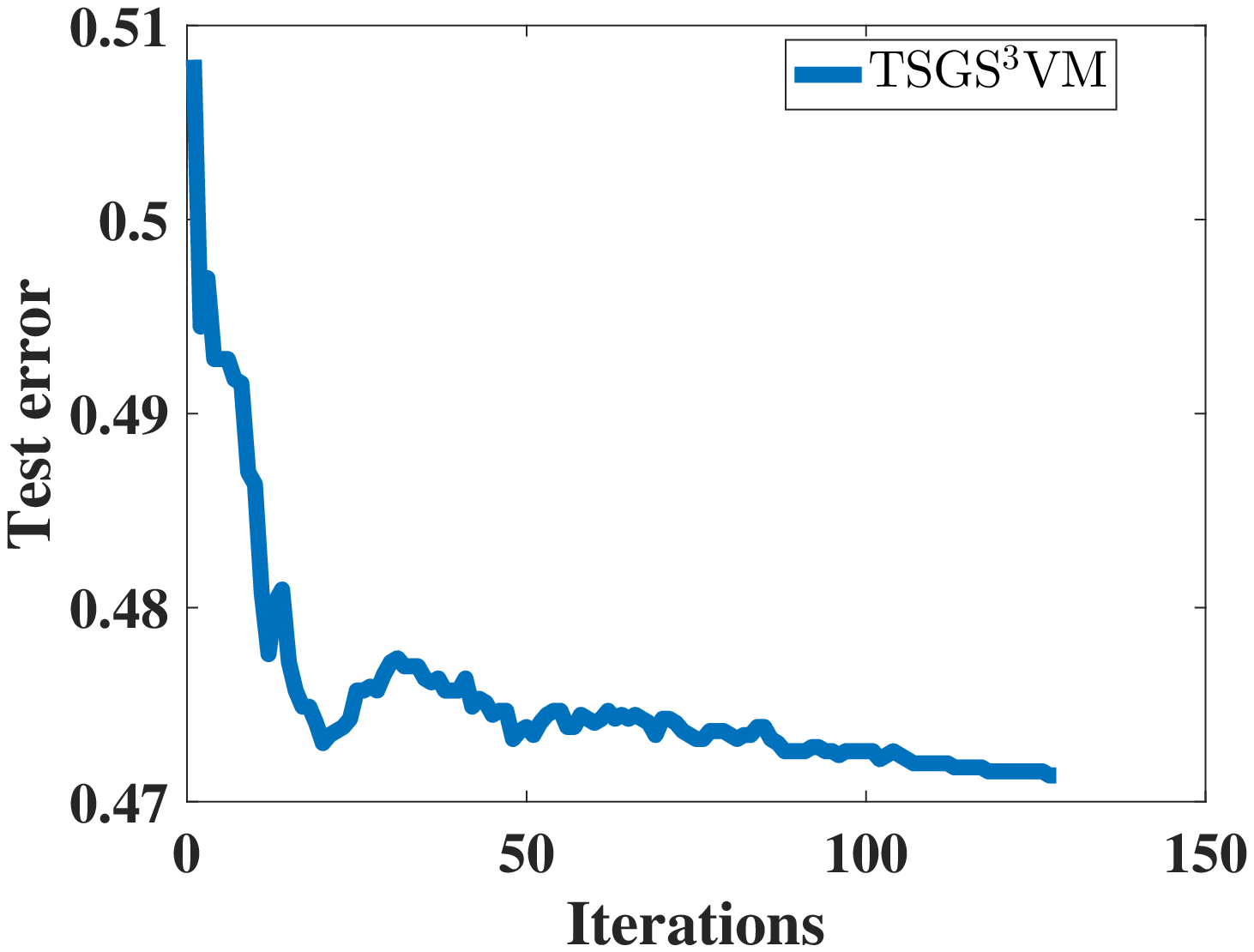}
		\caption{Dota2}
	\end{subfigure}
	\begin{subfigure}[]{0.24\textwidth}
		\includegraphics[width=1.6in]{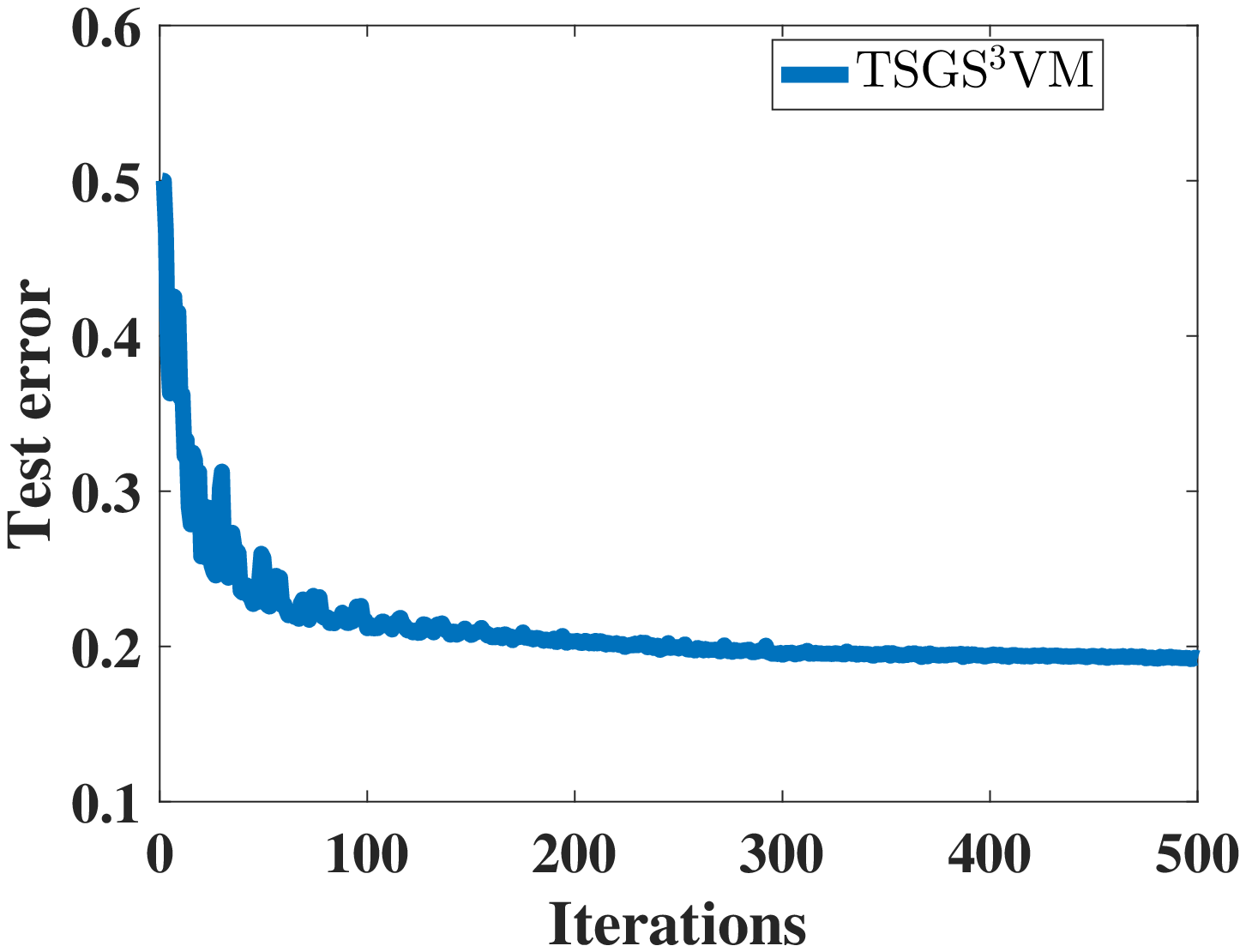}
		\caption{HEPMASS}
	\end{subfigure}
	\hspace{-10pt}
	\caption{Test error of different S$^3$VM solvers v.s. iterations on the eight  benchmark data sets.
	}
	\label{fig:error-iter}
\end{figure*}
\section{Convergence curves related to iterations}
We report the convergence curve related to iterations of TSGS$^3$VM in Fig \ref{fig:error-iter}. Fig \ref{fig:error-iter} shows that TSGS$^3$VM usually converge to a good result in a few iterations (about 64-128). Note that, similar to DSG, our TSGS$^3$VM implementation also uses a mini-batch setting, where the batch size is set to 256. Thus, in each iteration, TSGS$^3$VM randomly sample 256 instances to compute the TSG.

\bibliographystyle{named}
\bibliography{ijcai19}

\end{document}